\documentclass[10pt,twocolumn,letterpaper]{article}

\usepackage{iccv}
\usepackage{times}
\usepackage{graphicx}
\usepackage{amsmath}
\usepackage{amssymb}
\usepackage{multirow}
\usepackage{booktabs}
\usepackage{subfigure}

\usepackage[pagebackref=true,breaklinks=true,letterpaper=true,colorlinks,bookmarks=false]{hyperref}

\iccvfinalcopy 

\def\iccvPaperID{1428} 

\ificcvfinal\fi
\begin{document}

\title{Learning Filter Basis for Convolutional Neural Network Compression}

\author{Yawei Li$^1$\thanks{Equal contribution}~, Shuhang Gu$^1$\footnotemark[1]~, Luc Van Gool$^{1, 2}$, Radu Timofte$^1$\\
$^1$Computer Vision Lab, ETH Zurich, Switzerland, $^2$KU Leuven, Belgium\\
{\tt\small \{yawei.li, shuhang.gu, vangool, radu.timofte\}@vision.ee.ethz.ch}
}

\maketitle

\begin{abstract}
 Convolutional neural networks (CNNs) based solutions have achieved state-of-the-art performances for many computer vision tasks, including classification and super-resolution of images. Usually the success of these methods comes with a cost of millions of parameters due to stacking deep convolutional layers. Moreover, quite a large number of filters are also used for a single convolutional layer, which exaggerates the parameter burden of current methods. Thus, in this paper, we try to reduce the number of parameters of CNNs by learning a basis of the filters in convolutional layers. For the forward pass, the learned basis is used to approximate the original filters and then used as parameters for the convolutional layers. We validate our proposed solution for multiple CNN architectures on image classification and image super-resolution benchmarks and compare favorably to the existing state-of-the-art in terms of reduction of parameters and preservation of accuracy. Code is available at \url{https://github.com/ofsoundof/learning_filter_basis}.
\end{abstract}


\section{Introduction}
\label{sec:introdcution}

Recently, deep convolutional neural network (CNN) based approaches have been setting new state-of-the-art results not only for high-level computer vision tasks such as image classification~\cite{krizhevsky2012imagenet,simonyan2014very,he2016deep,huang2017densely}, segmentation~\cite{girshick2014rich,long2015fully}, and object detection~\cite{girshick2014rich,girshick2015fast,ren2015faster,redmon2016you}, but also for low-level tasks such as image super-resolution (SR)~\cite{dong2014learning,kim2015accurate,lim2017enhanced,li2018carn,zhang2019deep,li2019_3dappearance}, denoising~\cite{zhang2017beyond,zhang2017learning,gu2018multi}, and deburring~\cite{pan2016blind,lee2018joint}. However, most of the advances are achieved at the expense of relying on deeper architectures, millions of optimization parameters and resource-intensive computations. This hampers the application of deep neural networks under resource constrained environments, \eg, mobile phones.

To overcome the above mentioned problem, one research direction is to design efficient architectures. For example, in comparison with VGG~\cite{simonyan2014very}, both ResNet~\cite{he2016deep} and DenseNet~\cite{huang2017densely} reduce the number of parameters in their CNN by one magnitude of order while achieving comparable or even more accurate image classification results. Despite being more compact, there are still redundancies in those networks, making it possible for further compression.

In the meanwhile, network compression promises to alleviate the model complexity without losing much accuracy of the original network. Many network compression methods have been proposed. They mainly fall into three categories including network quantization~\cite{rastegari2016xnor,li2016ternary,courbariaux2016binarized,zhou2017incremental}, network pruning~\cite{han2015deep,han2015learning,he2017channel}, and filter decomposition~\cite{jaderberg2014speeding,zhang2016accelerating,wang2017factorized,son2018clustering,peng2018extreme}. In this paper, we focus on filter decomposition.

Filter decomposition approximates the original
filter with a lightweight convolution and a linear projection. Current methods either operate directly on the channel-wise 2D $w \times h$ filters~\cite{jaderberg2014speeding,wang2017factorized,son2018clustering} or decompose the intact 3D $c \times w \times h$ filters~\cite{zhang2016accelerating,peng2018extreme}. For those working on 2D filters, considering that the kernel size is usually small (\eg, $3 \times 3$) and a couple of new parameters are introduced to represent the 2D kernel, the compression ratio in terms of reduction of parameters is not impressive~\cite{wang2017factorized,son2018clustering}. The other filter decomposition methods~\cite{zhang2016accelerating} consider a 3D kernel as an intact element making impossible the reduction of the number of input channels~\cite{he2017channel}. This prevents the application of the method to narrow networks with much fewer output channels but more input channels. For example, in DenseNet-12-40, there are only 12 output channels which makes it not economic to decompose the 3D filters. 

The aforementioned methods either collapse or maintain the 3D filters during decomposition, which can be regarded as 'hard' decomposition. They are only coarse-grained configurations on the two boundary operating points. The motivation of this paper is to provide the missing in-between fine-grained operating points and to balance the parameter distribution between the two decomposed convolutions. Thus, we propose a novel filter basis learning method that circumvents the limitation of the 'hard' filter decomposition methods. We split the 3D filters along the input channel dimension and each split is considered as a basic element. We assume that the ensemble of those basic elements within one convolutional layer can be represented by the linear combinations of a basis. Our aim is to learn the basis and the linear combination together. During inference, the basis can be combined to reconstruct the original element, \ie, the 3D split. Then the splits are stacked along the input channel dimension to form the original 3D filters. Also, as it will be explained in the paper, the convolutions with respect to the original 3D filters can be converted to convolutions with respect to the learned basis. Thus, our method can be easily and efficiently implemented and embedded into the state-of-the-art networks.
Compared with previous works, our basis learning method also generalized easily to $1 \times 1$ convolution, which is vital for compressing networks with intensive $1 \times 1$ convolutions.

The contribution of this paper is four-fold. \textbf{(I)} We propose a \textit{novel basis learning method} that can reduce the input channel, making it eligible for narrow networks. Our method can be applied to convolutional layers with different kernel sizes and even $1 \times 1$ convolutions. \textbf{(II)} Our method achieves \textit{state-of-the-art compression} performance. On VGG, SRResNet, and EDSR, our method outperforms state-of-the-art compression method gracefully with lower classification error and fewer parameters of the previous compression method~\cite{son2018clustering,wang2017factorized}. On ResNet, DenseNet, our compressed model is comparable with the recent state-of-the-art with fewer parameters. \textbf{(III)} Our method generalizes easily to prior work just by changing the number of splits, thus leading to \textit{a unified formulation} of different filter decomposition methods~\cite{jaderberg2014speeding,zhang2016accelerating,wang2017factorized}. \textbf{(IV)} We validate our method on both high-level vision tasks, \ie, image classification, and \textit{low-level vision tasks}, \ie, \textit{image SR}. Compared with the high-level vision tasks such as image classification where a single class is regressed, the low-level image SR task is more challenging since the algorithm need to recover every pixel and the content detail in the image. However, none of the previous works apply compression method to networks designed for low-level task. Our experimental results show that network compression method also works well for image SR.

The reminder of the paper is organized as follows. In Sec.~\ref{sec:related_work}, we discuss the relevant work. In Sec.~\ref{sec:method}, we introduce the proposed filter basis learning method for network compression in detail. In Sec.~\ref{sec:learning_filter_basis}, we explain how to learn the basis filter and the coding coefficients. In Sec.~\ref{sec:experimental_result}, we provide the implementation details of our method and discuss the experimental results. Sec.~\ref{sec:conclusion} concludes the paper.

\section{Related Work}
\label{sec:related_work}

Network compression as a research topic attracted an increased interest recently. The works in this field can be roughly grouped into three categories, namely, network pruning, network quantization, and filter decomposition. 

\textbf{Network Pruning:} Network pruning attempts to prune the less important network parameters in the network. Han~\etal~\cite{han2015deep,han2015learning} tried to learn sparse connections and prune the less important ones. They introduced deep compression that combines several techniques such as weight pruning, quantization and weight sharing, and Huffman coding to reduce the size of neural networks. However, their method results in irregular kernel shapes making it difficult for implementation although the theoretical speed-up ratio is impressive~\cite{han2015learning}. Thus, channel pruning is proposed to remove redundant channels in feature maps which result in regular kernel shapes and implementation-friendly algorithms~\cite{alvarez2016learning,zhou2016less,wen2016learning}. Wen~\etal~\cite{wen2016learning} explored structured sparsity including channel-wise, shape-wise, and depth-wise sparsity in deep neural networks. He~\etal~\cite{he2017channel} proposed channel pruning to accelerate deep neural networks. Their method can choose representative channels and prune redundant ones, based on LASSO regression. 

\textbf{Network Quantization:} Network quantization aims at reducing the model size of neural networks by quantizing the weight parameters. Han~\etal~\cite{han2015deep} demonstrated how to quantize weight parameters to a relatively small number of shared weights without loss of accuracy. Chen~\etal~\cite{chen2015compressing} introduced a hash function to group network connections into hash buckets and forced connections falling into the same buckets to share the same weight. Other works attempt to reduce the precision of parameter by introducing binary~\cite{rastegari2016xnor,courbariaux2016binarized,courbariaux2015binaryconnect} and ternary~\cite{zhu2016trained} weights.

\textbf{Filter Decomposition:} Apart from the two aforementioned methods, filter decomposition is proposed to approximate the original filter with parameter efficient representations~\cite{denton2014exploiting,jaderberg2014speeding,lebedev2014speeding,zhang2016accelerating,danelljan2017eco,wang2017factorized,peng2018extreme}. Early low-rank approximation applies matrix decomposition by using SVD~\cite{denton2014exploiting} or CP-decomposition~\cite{lebedev2014speeding}. Jaderberg~\etal~\cite{jaderberg2014speeding} proposed to approximate the 2D filter set by a linear combination of a smaller basis set of 2D separable filters. Wang~\etal~\cite{wang2017factorized} built on the work of Jaderberg~\etal and further rearranged the decomposed filter sequentially. In their work, each normal convolution is decomposed into several layers of depth-wise convolution followed by $1 \times 1$ convolution. Son~\etal~\cite{son2018clustering} proposed to use k-means algorithm to cluster the $3 \times 3$ convolutional kernels. The kernels that fall in the same cluster share the same weight parameter. However, for each $3 \times 3$ kernel, a scale and an index parameter is introduced to represent the kernel. So the compression ratio in terms of number of parameters is fixed and slightly larger than 2/9. The same problem exists for~\cite{wang2017factorized}. Although the compression ratio 1/9 could be achieved by~\cite{wang2017factorized}, the classification accuracy is severely diminished. Another drawback of~\cite{son2018clustering} is that it could not be applied to $1 \times 1$ convolutions favored by modern networks such as ResNet and DenseNet.

Instead of working on 2D filters as in the previous low-rank approximation, Zhang~\etal~\cite{zhang2016accelerating} directly dealt with 3D filters by considering the input channel as the third dimension. However, their method cannot reduce the input channel. This prohibits the application of the decomposition method narrow networks with small output channel but large input channel such as DenseNet. In a recent work, Peng~\etal~\cite{peng2018extreme} proposed to approximate a normal convolution by group convolution followed by a linear combination ($1 \times 1$ convolution). However, they did not apply their approximation methods to DenseNet, which is of particular interest in the newly proposed architectures.

By contrast, our proposed basis learning method can be applied to convolutions with any kernel size and any input/output channel size. This makes our method flexible to compress different modern networks.

\section{Filter Decomposition for Network Compression}
\label{sec:method}

Given an input image $x \in \mathcal{X}$, the aim of supervised learning is to recover the corresponding label $y \in \mathcal{Y}$. For low-level vision tasks such as image SR, the label is the ground-truth high-resolution image corresponding to the low-resolution input image $x$. For high-level image classification, $y$ is a class label of the image. The regression process can be represented by a simple function
\begin{equation}
    \hat{y} = f_{\Theta}(x), 
    \label{eqn:regression}
\end{equation}
where $\hat{y}$ denotes the regressed label and $f_{\Theta}(\cdot)$ is the regression function of the neural network parameterized by $\Theta$. 

\subsection{Decomposing convolution layer with filter basis}
\label{subsec:decomposing_convolution_layer}

We assume that a convolution layer has $c$ input channels and $n$ output channels, and the kernel size is $w\times h$.
In order to reduce the number of parameters in neural network, different decomposition methods have been suggested. 
Zhang \etal assumed the parameters of a convolution layer could be approximated by a low-rank matrix~\cite{zhang2016accelerating}, \ie, 
\begin{equation}
    \mathbf{W} \approx \mathbf{B} \cdot \mathbf{A},
 \label{eqn:reconstructionmatrix}
\end{equation}
\begin{align}
    \mathbf{W} \approx \mathbf{B} \cdot \mathbf{A} \\
    \mathbf{W}\in\Re^{{cwh} \times n} = [\mathbf{W}_1, \cdots, \mathbf{W}_n] \\
    \mathbf{W}\in\Re^{{wh} \times cn}
 \label{eqn:reconstructionmatrix}
\end{align}
where $\mathbf{W}\in\Re^{{cwh} \times n} = [\mathbf{W}_1, \cdots, \mathbf{W}_n]$ is the matrix that contains the vectorized 3D filters, the multiplication of matrix $\mathbf{B}\in\Re^{{cwh}\times m}$ and matrix $\mathbf{A}\in\Re^{{m}\times n}$ is a low-rank matrix with rank $m<n$.
Besides formulating the parameters of convolution layer as a ${cwh} \times n$ matrix, there are also other low-rank approximation works \cite{jaderberg2014speeding,wang2017factorized} which consider the parameter matrix as a ${wh}\times {cn}$ matrix.
These works treat each channel in the 3D filter independently. 

\begin{figure}[t]
    \begin{center}
        \includegraphics[width=0.8\linewidth]{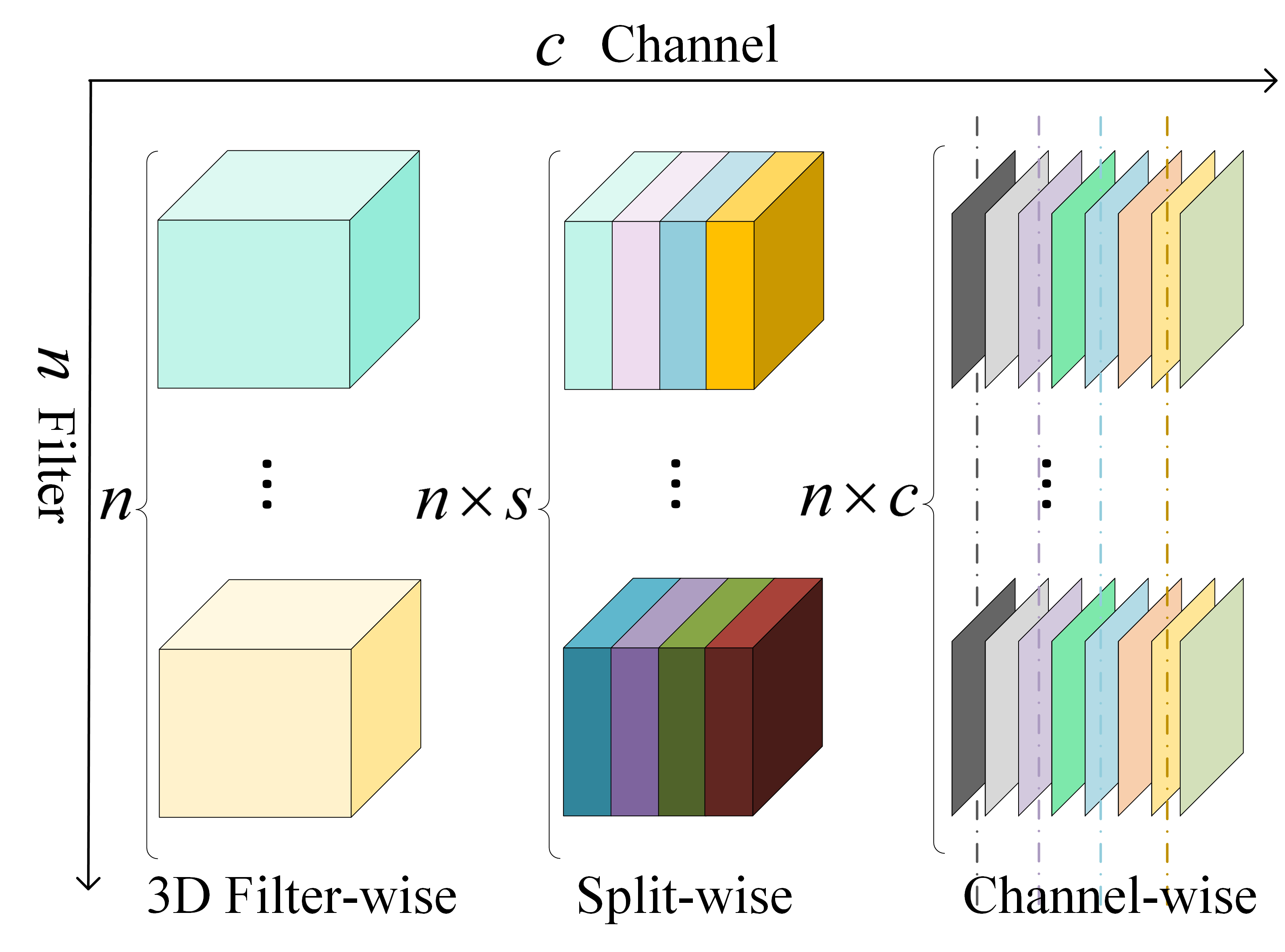}
    \end{center}
    \vspace{-0.3cm}
    \caption{Comparison of different filter decomposition methods. \textit{Right:} each channel of the 3D filter is considered as a basic element. A unique set of basis is learned for the $n$ 2D filters in each channel. \textit{Middle (the proposed):} the 3D filters are split into $s$ groups along the channel dimension and each group is considered as a basic filter element. A basis set is learned for all of the $n \times s$ splits of all the 3D filters. \textit{Left:} the 3D filter is considered as a whole. A basis set is learned for the 3D filters.}
    \vspace{-0.4cm}
\label{fig:filter_reduction}
\end{figure}

In Eqn.~\eqref{eqn:reconstructionmatrix}, the approximation of filter can also be analyzed in a filter basis decompostion perspective. Each 3D filter $\mathbf{W}_i\in\Re^{{cwh}\times 1}$ (or $\mathbf{W}_i\in\Re^{{wh}\times 1}$ for the channel-wise decomposition case) is represented by the linear combination of a set of $m$ filter basis $\{\mathbf{B}_j | j = 1, \cdots, m \}$ with the coding coefficient vector $\mathbf{A}_{i}\in\Re^{m\times1}$:
\begin{equation}
    \mathbf{W}_i \approx \sum_{j=1}^{m} {\alpha_{j,i}\mathbf{B}_j}, i = 1, \cdots, n.
    \label{eqn:reconstruction}
\end{equation}
where $\mathbf{A}_{i}$ is the $i$-th column of $\mathbf{A}$, $\mathbf{B}_j$ is the $j$-th filter basis with dimension $cwh \times 1$ or $wh \times 1$ for the 3D filter-wise decomposition and 2D channel-wise decomposition cases, respectively.
An illustration of direct 3D filter-wise decomposition and channel-wise filter decomposition can be found in the left and right part of Fig.~\ref{fig:filter_reduction}.

From the viewpoint of filter basis decomposition,  more flexible decomposition strategy can be adopted.
In the next subsection, we analyze the relationship between the dimension of filter basis and compression rate, and suggest a split-wise decomposition approach for network compression.

\subsection{Compression rate with different filter basis}
\label{subsec:compression_rate}

If we utilize $m$ 3D filter basis as a basic element (Fig~\ref{fig:filter_reduction}: \textit{Left}) to decompose the parameters of a convolution layer, the compression rate of the parameters is
\begin{equation}
    \Gamma_{filter} = \frac{m \cdot c \cdot w \cdot h + m \cdot n}{n \cdot c \cdot w \cdot h} = \frac{m}{n} + \frac{m}{c \cdot w \cdot h},
    \label{eqn:reduction_rate1}
\end{equation}
where $(n \cdot c \cdot w \cdot h)$, $(m \cdot c \cdot w \cdot h)$, and $(m \cdot n)$ is the number of parameters of the original convolution layer, the filter basis, and the coding coefficients, respectively.
In most of existing neural networks, $c \cdot w \cdot h$ is much larger than $n$. Thus, the first term in Eqn.~\eqref{eqn:reduction_rate1} dominates the compression rate.
For the 2D channel-wise decomposition case, we can similarly get the compression rate, namely,
\begin{equation}
    \Gamma_{channel} = \frac{m  \cdot w \cdot h + c \cdot m \cdot n}{n \cdot c \cdot w \cdot h} = \frac{m}{n \cdot c} + \frac{m}{w \cdot h}.
    \label{eqn:reduction_rate2}
\end{equation}
The major storage budget is used for the coding coefficients.
\begin{equation}
    \Gamma_{filter} = \frac{m}{n} + \frac{m}{c \cdot w \cdot h}
    \label{eqn:reduction_rate1}
\end{equation}
\begin{equation}
    \Gamma_{channel} = \frac{m}{n \cdot c} + \frac{m}{w \cdot h}
    \label{eqn:reduction_rate2}
\end{equation}
\begin{equation}
    \Gamma_{split} = \frac{m}{n \times s} + \frac{m}{p \cdot w \cdot h}
    \label{eqn:reduction_rate3}
\end{equation}

In order to achieve a better trade-off between compressing the basis and coefficients, we split the 3D filters along the channel dimension as illustrated in the middle part of Fig.~\ref{fig:filter_reduction}, namely, thinking of the $c \times w \times h$ filter as being composed of $s$ smaller $p \times w \times h$ filters and $c=s \cdot p$. 
As a result, the $n$ 3D $c \times w \times h$ filters  can be regarded as $n \cdot s$ filters with size $p \times w \times h$. Then, the problem becomes learning the basis and the representation coefficients of the $n \cdot s$ smaller filters. And the compression rate becomes
\begin{equation}
    \Gamma_{split} = \frac{m \cdot p \cdot w \cdot h + m \cdot n \cdot s}{n \cdot c \cdot w \cdot h} = \frac{m}{n \times s} + \frac{m}{p \cdot w \cdot h}.
    \label{eqn:reduction_rate3}
\end{equation}
The compression rate equation in Eqn.~\eqref{eqn:reduction_rate3} enable us to utilize generalized split-wise decomposition formula to achieve better compression rate.
Concretely, the optimal compression rate with respect to the size of filter basis could be achieved by solving the following optimization problem:
\begin{align}
    \{s^*, p^*\} &= \underset{\{s, p\}}{\arg\min} \left \{\frac{m}{n \times s} + \frac{m}{p \cdot w \cdot h} \right \} \quad \text{s.t.} \quad c = s \cdot p \nonumber \\
    &= \left\{\sqrt{\frac{c \cdot w \cdot h}{n}}, \sqrt{\frac{n \cdot c}{w \cdot h}}\right\}.
\end{align}
We can further quantize $p$ to the nearest integer that can divide $c$. For most of the convolutional layers, the input channel $c$ and output channel $n$ are the same or of the same magnitude order, \ie, $c \approx n$. Thus, the optimal group $s^* \approx \sqrt{w \times h}$. That is to say, the optimal configuration of splits is neither Fig.~\ref{fig:filter_reduction}:\textit{~Left} nor Fig.~\ref{fig:filter_reduction}:\textit{~Right} but the middle state between them.

\subsection{Implementing with convolution}
\label{subsec:implement_with_convolution}

\begin{figure}[t]
    \begin{center}
       \includegraphics[width=0.8\linewidth]{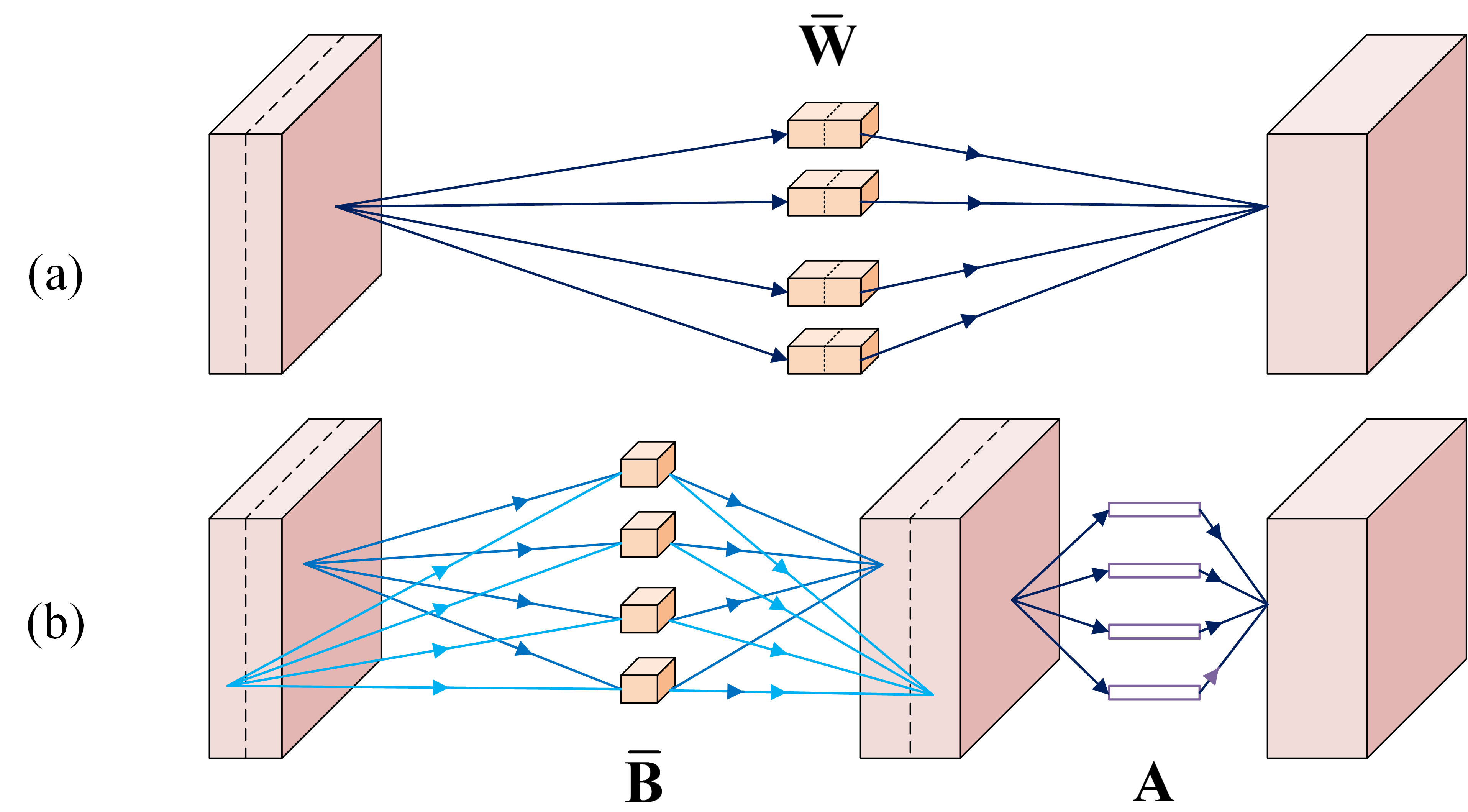}
    \end{center}
    \vspace{-0.3cm}
   \caption{Illustration of the proposed basis learning method. Operations are converted to convolutions. Unlike the normal convolution, our method splits both the input feature map and the 3D filter along the channel dimension. A set of basis is learned for the ensemble of splits. Every split of the input feature map is convolved with the basis. A final $1 \times 1$ convolution generates the output.}
   \vspace{-0.4cm}
\label{fig:convert_to_convolution}
\end{figure}

In this subsection, we show that filter decomposition can be implemented by convolution in the forward pass. By rearranging the operation, the proposed filter decomposition approach can alleviate the computation burden and compress network parameters.

We start with the case where there is only one split, \ie, $s=1$. 
As in Eqn.~\ref{eqn:reconstruction}, we utilize linear combination of filter basis to reconstruct the 3D filter $\mathbf{W}_i = \sum_{j=1}^{m} {\alpha_{j,i}\mathbf{B}_j}$. For the simplicity of notation, we use the same notation to represent the original non-vectorized 3D filters and basis, \ie, $\mathbf{W}_i, \mathbf{B}_j\in\Re^{c \times w \times h}$. Thus, the convolution between the input feature map $x$ and the 3D kernel becomes
\begin{equation}
    x * \mathbf{W}_i = x * \left( \sum_{j=1}^{m} {\alpha_{j,i}\mathbf{B}}_j \right) = \sum_{j=1}^{m} {\alpha_{j,i} \left(x * \mathbf{B}_j \right) }.
    \label{eqn:convolution_decompose}
\end{equation}
The second equality follows the linearity of convolution. 
Eqn.~\eqref{eqn:convolution_decompose} decompose the convolution operation with 3D filter $\mathbf{W}_i$ as linear combination of convolution operations with filter basis $\{\mathbf{B}_j,j=1,\dots,m\}$.
The linear combination can be implemented by a $1 \times 1$ convolution. 

For more general split-wise decomposition cases, we use smaller filter basis $\{\overline{\mathbf{B}}_j\in\Re^{p \times w \times h},j=1,\dots,m\}$ to reconstruct each sub-part of the 3D filter, namely,
\begin{align}
    \mathbf{W}_i & = \Big[\overline{\mathbf{W}}_{i,1};\dots;\overline{\mathbf{W}}_{i,s} \Big], \\
    \overline{\mathbf{W}}_{i,g} & = \sum_{j=1}^{m} {\alpha_{j,i, g}\overline{\mathbf{B}}_j},
    \label{eqn:reconstruction_3d_group}
\end{align}
where $ \overline{\mathbf{W}}_{i,g}\in\Re^{p \times w \times h}$ is a split of the 3D filter, $g=1,\dots,s$ is the split index, and $[\cdot]$ is the operator that stacks the basis along the channel dimension. Accordingly, the convolution between $x$ and $\overline{\mathbf{W}}_i$ becomes 
\begin{align}
    x * \mathbf{W}_i &= \left[ \overline{x}_1, \cdots, \overline{x}_s \right] * \left[ \overline{\mathbf{W}}_{i,1}, \cdots, \overline{\mathbf{W}}_{i,s}  \right] \nonumber \\
    &= \sum_{g=1}^{s}{\overline{x}_g * \overline{\mathbf{W}}_{i,g}}
    = \sum_{g=1}^{s}{\overline{x}_g * \sum_{j=1}^{m} {\alpha  _{j,i, g}\overline{\mathbf{B}}_j}} \nonumber \\
    &= \sum_{g=1}^{s} \sum_{j=1}^{m} {\alpha_{j,i, g} \left( {\overline{x}_g *  \overline{\mathbf{B}}_j} \right) }.
    \label{eqn:convolution_decompose_group}
\end{align}
where $\{\overline{x}_g, g=1,\dots,s\}$ in Eqn.~\eqref{eqn:convolution_decompose_group} are the splits of the input $x$.
As revealed by Eqn.~\eqref{eqn:convolution_decompose_group}, in the split-wise decomposition case, each split of feature map is firstly convolved with the filter basis, and then the final output is achieved by a weighted summation of the convolution results.
This operation on feature map splits could be implemented as a 3D convolution as in Pytorch~\cite{paszke2017automatic} or Tensorflow~\cite{abadi2016tensorflow} with stride $p=c/s$ and no padding along the channel dimension. 
But we find the 3D convolution implementation is not efficient. In this way it takes 121 ms to run the compressed EDSR model for one iteration with batch size 16 and patch size $48 \times 48$. Instead, we implement the operation with $s$ 2D convolutions that share the same weight parameter and the running time drops to 62 ms. 
The linear combination is again converted to a $1 \times 1$ convolution.
Thus, no matter how many splits there are, a standard convolution can be decomposed into a convolution with respect to the basis and a $1 \times 1$ convolution. The implementation is illustrated in Fig.~\ref{fig:convert_to_convolution}.

\subsection{Filter basis decomposition for special filter sizes}
\label{subsec:special_filter_sizes}

As shown in the above analysis, our basis learning method follows a general setting of filter size, \ie, $n \times c \times w \times h$. This means that the proposed basis learning method can be applied to any convolutional filters. Here we emphasize two special filter sizes. 

\textbf{$1 \times 1 $ convolution:} The first one is $1 \times 1 $ convolution which is favored by modern neural networks~\cite{he2016deep,huang2017densely}. When the input/output channels are quite large, considerable parameters and computation are consumed by $1 \times 1 $ convolution. For example, in DenseNet-12-40 architecture~\cite{huang2017densely}, 12.1\% of the parameters is in the two large $1 \times 1 $ convolutions. Unfortunately, prior filter decomposition works~\cite{jaderberg2014speeding,denton2014exploiting,wang2017factorized} could not be applied to this kind of convolution. Following our formulation Eqn.~\eqref{eqn:convolution_decompose} through Eqn.~\eqref{eqn:convolution_decompose_group}, a $1 \times 1 $ convolution with large $n$ and $c$ can be decomposed into two cheaper ones. 

\textbf{$c \gg n > m$ convolution:} In some networks such as DenseNet, the output channel $n$ is much smaller than the input channel $c$. In this case, according to Eqn.~\eqref{eqn:reduction_rate1}, we are in the dilemma of either choosing a even smaller basis size $m$ at the risk of losing too much accuracy or selecting an $m$ comparable with $n$ thus resulting in uneconomic compression. As revealed by Eqn.~\eqref{eqn:reduction_rate2}, by splitting the 3D kernels along the channel dimension, we can have $s$ times more filters. 
So we can gracefully choose a comfortable basis size that leads to both economic compression and high accuracy. 

\section{Learning Filter Basis}
\label{sec:learning_filter_basis}
In the previous section, we have shown that decomposing filter splits into linear combinations of filter basis could reduce the computational burden and parameter number of networks.
In this section, we present our learning method for learning filter basis.

\subsection{General filter basis learning approach}
\label{subsec:learning_approach}

For the purpose of notation simplicity, we only introduce the simple case of using $\mathbf{B}\cdot\mathbf{A}$ to approximate filter $\mathbf{W}$.
The training method for the split-wise case of approximating $\overline{\mathbf{W}}$ with $\overline{\mathbf{B}}\cdot\overline{\mathbf{A}}$ is exactly the same.
We jointly minimize the approximation error $\|\mathbf{W}-\mathbf{B}\cdot\mathbf{A}\|_F^2$ and the network target loss $\mathcal{L}(y, f{(x)})$.
For example, to compress image restoration network with mean square error (MSE) loss, our training objective function is
\begin{equation}
\min_{\mathbf{B}^l, \mathbf{A}^l} \left \| y - f_{{\mathbf{B},\mathbf{A}}|\Theta}(x)\right \|^2_F  + \gamma \sum_{l=1}^{L} \left \| \mathbf{W}^l - {\mathbf{B}^l \cdot \mathbf{A}^l} \right \|_F^2,
\label{eqn:loss}
\end{equation}
where $f_{{\mathbf{B},\mathbf{A}}|\Theta}(\cdot)$ denotes the CNN with parameter $\{\mathbf{B},\mathbf{A}\}$ conditioned that the other parameters $\Theta$ are known and the superscript $l$ indexes the $l$-th layer of an $L$-layer network.

After having learned the basis and the coding matrices $\{\mathbf{B},\mathbf{A}\}$, there is no need to store the original filters. During inference, $\{\mathbf{B},\mathbf{A}\}$ is used as the weight parameter as the lightweight and $1 \ \times 1$ convolution, respectively. The total number of parameters of the basis and the coding coefficient is much fewer than those of the original filters, thus achieving a reduction of the number of parameters.

\begin{figure}[t]
\begin{center}
   \includegraphics[width=0.9\linewidth]{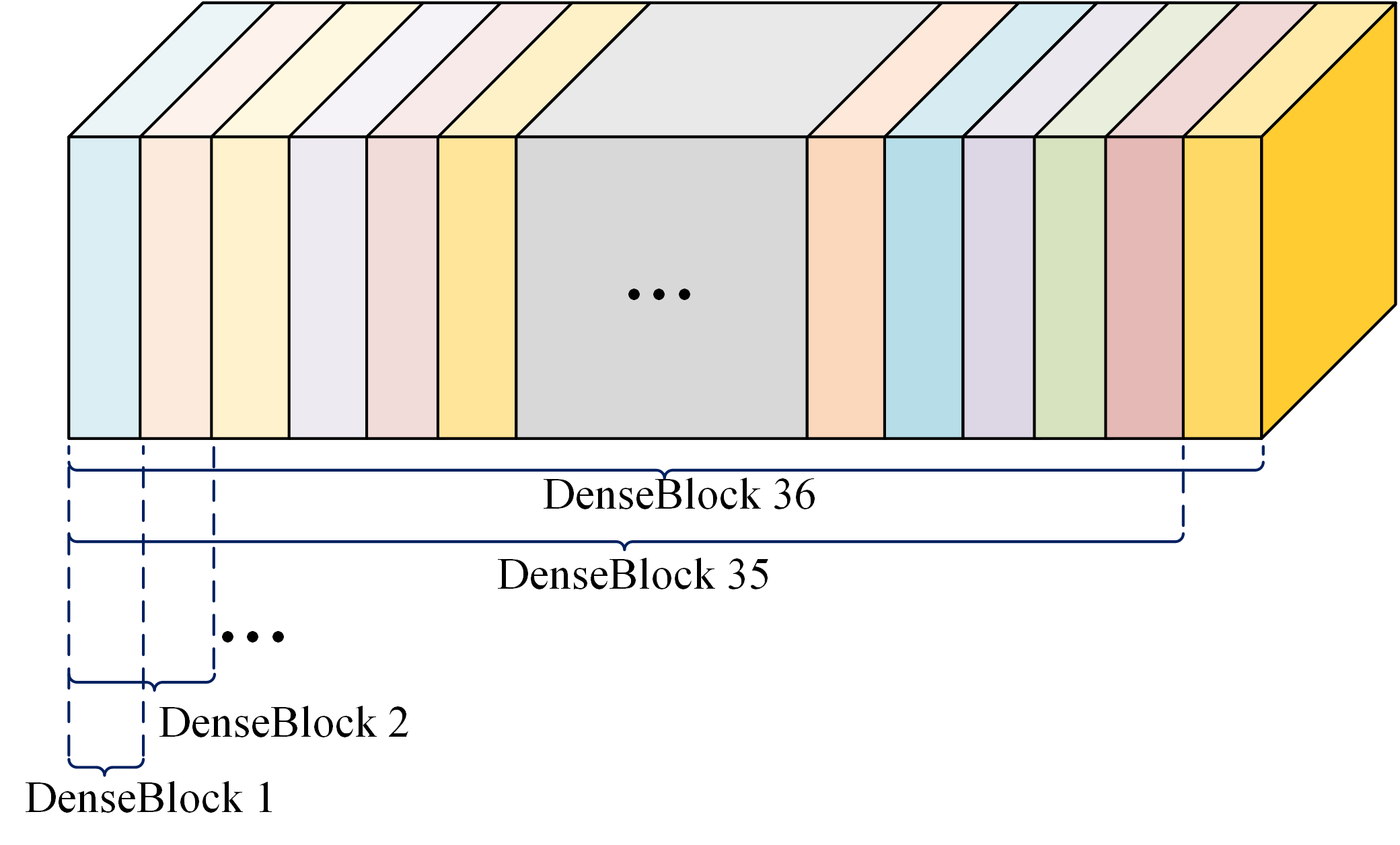}
\end{center}
\vspace{-0.2cm}
   \caption{Basis sharing for the compression of DenseNet-12-40. The basis set is shared by all of the DenseBlocks in DenseNet-12-40. The shared basis is split into 36 splits. The basis of a certain DenseBlock is sliced from the shared basis set. Starting from the lower DenseBlock, every DenseBlock adds a new split from the shared basis set to the basis of the previous block forming the basis of current block. Thus, the basis channel of the DenseBlock grows gradually.}
\label{fig:basis_sharing_densenet}
\vspace{-0.4cm}
\end{figure}

\subsection{Basis sharing}
\label{subsec:basis_sharing}

To compress the networks further, we can force several or all convolutional layers to share the same basis set depending on the compression degree we want to achieve. The weight sharing strategy can be customized to the networks. For example, in ResNet~\cite{he2016deep} and the following works SRResNet~\cite{ledig2016photo}, EDSR~\cite{lim2017enhanced}, there are two convolutions in the residual block. We can let the two convolutions share the basis. In ResNet-56 architecture for CIFAR10, the residual blocks are grouped into three groups, each with 9 residual blocks and increasing feature map channels. The channels in the lower residual block groups are relatively small (16 and 32 for the first and second group). To achieve a satisfying compression rate, we have the convolutions within the same group share a common basis set. Moreover, in DenseNet, the input channel grows gradually with a step 12. There is no clear sign like in ResNet to indicate which convolution should share the basis. In this case, all of the convolutional layers share the same basis. For the lower layers, only a slice of the basis is used while only for the last convolutional layer the whole basis is used. The basis sharing strategy for DenseNet-12-40 is shown in Fig.~\ref{fig:basis_sharing_densenet}.

In conclusion, we can  apply block-wise, group-wise, or network-wise basis sharing flexibly according to the architecture of the target network. 

\begin{table}[t]
    \footnotesize
    \begin{center}
        \begin{tabular}{c|c|c|c|c}
            \toprule
            \multirow{3}{*}{Metrics} & \multicolumn{3}{c|}{Basis Share} &\multirow{3}{*}{\shortstack{Base-\\line}}\\ \cline{2-4}
            & No / Yes & No /Yes & No / Yes \\ 
            & $m=16$ &$m=32$ &$m=64$ \\ \midrule
            Set5        &32.14 / 32.16 &32.22 / 32.20 &32.33 / 32.30 &32.48 \\ 
            Set14       &28.58 / 28.57 &28.66 / 28.64 &28.72 / 28.73 &28.81\\ 
            B100        &27.58 / 27.57 &27.62 / 27.61 &27.66 / 27.64 &27.72\\ 
            Urban100    &26.05 / 26.00 &26.20 / 26.20 &26.38 / 26.38 &26.65 \\
            DIV2K       &28.96 / 28.93 &29.06 / 29.04 &29.14 / 29.14 &29.25\\ \hline
            \#Params & 27k / 17k &53k / 35k &106k / 70k &1180k \\ \hline
            Comp. (\%) & 2.3 / 1.5 &4.5 / 3.0 &9.0 / 5.9 &100\\
            \bottomrule
        \end{tabular}
    \end{center}
    \vspace{-0.1cm}
    \caption{Different operating points of applying the proposed basis learning to EDSR for image SR for upscaling factor $\times 4$. PSNR (dB) is reported for the five commonly used datasets. `Basis Share' indicates whether the two convolutions within the same residual block share the basis. $m$ is the number of basis. The number of splits $p$ for one convolution is 4.}
    \label{tbl:super_resolution_edsr_ablation}
    \vspace{-0.4cm}
\end{table}

\begin{table*}[t]
    \small
    \begin{center}
        \begin{tabular}{c|c|c|c|c|c|c}
            \toprule
            \multicolumn{2}{c|}{SRResNet~\cite{ledig2016photo}} & Factor-SIC2 &Factor-SIC3 & Basis-64-14 (ours) & Basis-32-32 (ours) & Baseline \\ \midrule
            \multirow{4}{*}{PSNR (dB)} & Set5 &31.68 &31.86 &31.84 &31.90 &32.03  \\ 
            & Set14    &28.32 &28.38 &28.38 &28.43 &28.50  \\ 
            & B100     &27.37 &27.40 &27.39 &27.44 &27.52  \\ 
            & Urban100 &25.47 &25.58 &25.54 &25.65 &25.88  \\ 
            & DIV2K     &28.59 &28.65 &28.63 &28.69 &28.85  \\ \hline 
            \multicolumn{2}{c|}{\#Parameters}    &19k &28k &18k &27k &74k  \\ \hline
            \multicolumn{2}{c|}{Compression Rate (\%)} &25.3 &38.0 &24.3 &36.1 &100  \\ \midrule \midrule
            \multicolumn{2}{c|}{EDSR-8-128~\cite{lim2017enhanced}} & Factor-SIC2 &Factor-SIC3 & Basis-128-27 (ours) & Basis-128-40 (ours) & Baseline \\ \midrule
            \multirow{4}{*}{PSNR (dB)} & Set5 &31.82 &31.96 &31.95 &32.03 &32.10  \\ 
            & Set14    &28.40 &28.47 &28.42 &28.45 &28.55 \\ 
            & B100     &27.43 &27.49 &27.46 &27.50 &27.55 \\ 
            & Urban100 &25.63 &25.81 &25.76 &25.81 &26.02 \\ 
            & DIV2K     &28.70 & 28.81 &28.76 &28.82 &28.93 \\ \hline
            \multicolumn{2}{c|}{\#Parameters}    &70k &105k &69k &102k &295k \\ \hline
            \multicolumn{2}{c|}{Compression Rate (\%)} &23.8 &35.7 &23.4 &34.7 & 100  \\
            \bottomrule
        \end{tabular}
    \end{center}
    \vspace{-0.2cm}
    \caption{Comparison of Factor~\cite{wang2017factorized} and our basis learning method for the lighter SR networks SRResNet and EDSR-8-128. The upscaling factor is $\times 4$. For each method, two operating points including a lighter one and a heavier one are reported. `SIC*' denotes the number of SIC layers in Factor. `Basis-N-S' means that the number of basis is S and each basis has N input channels.}
    \label{tbl:super_resolution_srresnet_edsr}
    \vspace{-0.2cm}
\end{table*}

\begin{table*}[t]
    \small
    \begin{center}
        \begin{tabular}{c|c|ccc|ccc|ccc|ccc}
            \toprule
            \multicolumn{2}{c|}{\multirow{2}{*}{Comparison Metrics}}  & \multicolumn{3}{c|}{Factor~\cite{wang2017factorized}} & \multicolumn{3}{c|}{\textbf{Basis-S (ours)}} & \multicolumn{3}{c|}{\textbf{Basis (ours)}} & \multicolumn{3}{c}{Baseline} \\ \cline{3-14}
            \multicolumn{2}{c|}{} & $\times2$ &$\times3$ &$\times4$& $\times2$ &$\times3$ &$\times4$ & $\times2$ &$\times3$ &$\times4$ & $\times2$ &$\times3$ &$\times4$ \\ \midrule
            \multirow{4}{*}{\shortstack{PSNR\\(dB)}} & Set5 &37.95 &34.33 &32.05 &38.09  &34.47 &32.29 &38.12 &34.55 &32.39 &38.19 &34.68 & 32.48   \\ \cline{2-14} 
            & Set14 &33.53 &30.31 &28.54 &33.75  &30.41 &28.63 &33.72 &30.46 &28.69 &33.95 &30.53 &28.81 \\ \cline{2-14} 
            & B100  &32.15 &29.08 &27.55 &32.23  &29.15 &27.62 &32.27 &29.18 &27.64 &32.35 &29.26 &27.72 \\ \cline{2-14} 
            & Urban100 &31.99 &28.10 &25.98 &32.38  &28.39 &26.25 &32.46 &28.51 &26.36 &32.97 &28.81 &26.65 \\ \cline{2-14}
            & DIV2K &34.60 &30.91 &28.92 &34.77 &31.06 &29.06 &34.84 &31.11 &29.13 &35.03 &31.26 & 29.25 \\ \hline
            \multicolumn{2}{c|}{\#Parameters} &\multicolumn{3}{c|}{136k} &\multicolumn{3}{c|}{\textbf{90k}} &\multicolumn{3}{c|}{\textbf{164k}} &\multicolumn{3}{c}{1180k}  \\ \hline
            \multicolumn{2}{c|}{Compression Ratio (\%)} &\multicolumn{3}{c|}{11.5} &\multicolumn{3}{c|}{\textbf{7.6}} &\multicolumn{3}{c|}{\textbf{13.9}} &\multicolumn{3}{c}{100}  \\
            \bottomrule
        \end{tabular}
    \end{center}
    \vspace{-0.2cm}
    \caption{Compression results for EDSR~\cite{lim2017enhanced}. Basis-S uses basis sharing for the two convolutions within the same residual block.}
    \label{tbl:super_resolution_edsr}
    \vspace{-0.4cm}
\end{table*}

\begin{figure*}
\centering
\subfigure{
  \begin{minipage}[t]{0.195\linewidth}
    \centering
    \includegraphics[width=1\textwidth]{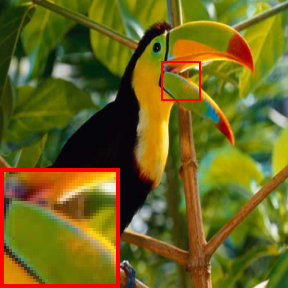}
    {\footnotesize  Ground Truth: PSNR (dB)}
  \end{minipage}

  \begin{minipage}[t]{0.195\linewidth}
    \centering
    \includegraphics[width=1\textwidth]{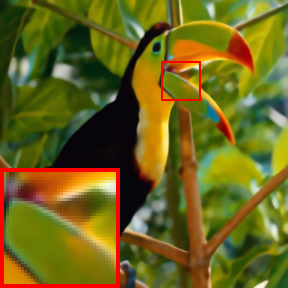}
    {\footnotesize Factor~\cite{wang2017factorized}: 34.67
    }
  \end{minipage}

  \begin{minipage}[t]{0.195\linewidth}
    \centering
    \includegraphics[width=1\textwidth]{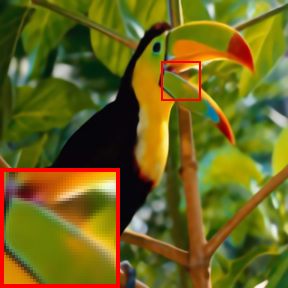}
    {\footnotesize Basis-S (ours): 34.99
    }
  \end{minipage}
  
  \begin{minipage}[t]{0.195\linewidth}
    \centering
    \includegraphics[width=1\textwidth]{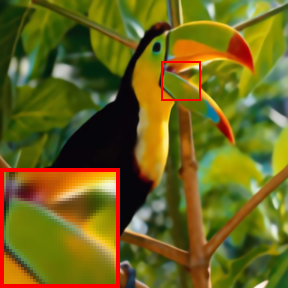}
    {\footnotesize Basis (ours): 35.10}
  \end{minipage}

  \begin{minipage}[t]{0.195\linewidth}
    \centering
    \includegraphics[width=1\textwidth]{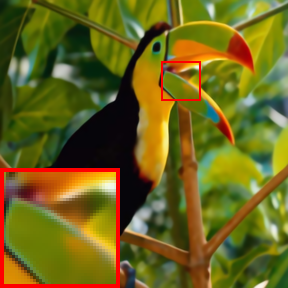}
    {\footnotesize Baseline: 35.27}
  \end{minipage}
}
\vspace{-0.15cm}
\caption{SR results of \textit{bird} image for upscaling factor $\times 4$. Network compression methods are applied on EDSR~\cite{lim2017enhanced}.} 
\label{fig:sr_bird}
\vspace{-0.2cm}
\end{figure*}

\section{Experimental Results}
\label{sec:experimental_result}

We show the experimental results in this section and compare with the state-of-the-art methods on both image classification and image SR. For classification, we applied our basis learning method to various networks including VGG~\cite{simonyan2014very}, ResNet~\cite{he2016deep}, and DenseNet~\cite{huang2017densely}. We evaluate the performance of compressed models on CIFAR10~\cite{krizhevsky2009learning} dataset. The training and testing subset contains 50,000 and 10,000 images, respectively. As is done by prior works~\cite{he2016deep,huang2017densely}, we normalize all images using channel-wise mean and standard deviation of the the training set. Standard data augmentation is also applied. We train the compressed networks for 300 epochs with SGD optimizer and an initial learning rate of 0.1. The learning rate is decayed by 10 after 50\% and 75\% of the epochs. 

For image SR, we applied our method to two typical SR networks, namely, SRResNet~\cite{ledig2016photo} and EDSR~\cite{lim2017enhanced}. SRResNet is a middle-level network with 1.5M parameters while EDSR is quite a huge network with 43M parameters but much higher PSNR accuracy. For fast training, we also compressed a lighter version of EDSR with 8 residual blocks and 128 channels per convolution in the residual block. We denote this network as EDSR-8-128. The networks are trained on DIV2K~\cite{Agustsson_2017_CVPR_Workshops} dataset that contains 1,000 2K images. We test the networks on five datasets: Set5~\cite{bevilacqua2012low}, Set14~\cite{zeyde2010single}, B100~\cite{MartinFTM01}, Urban100~\cite{Huang-CVPR-2015}, and DIV2K validation set. Adam optimizer~\cite{kingma2014adam} is used for training SR networks. We use the default hyper-parameter. The networks are trained for 300 epochs. The learning rate starts from $1\times 10^{-4}$ and decays by 10 after 200 epochs.

\subsection{Validation on super-resolution}
\label{subsec:super_resolution}

The compression results of SR networks are shown in Table~\ref{tbl:super_resolution_edsr_ablation}, Table~\ref{tbl:super_resolution_srresnet_edsr}, and Table~\ref{tbl:super_resolution_edsr}. In Table~\ref{tbl:super_resolution_edsr_ablation}, we explore different operating points applied to EDSR. We use 4 splits for the convolution in the residual block. For a clearer comparison, we report the number of parameters and compression ratio for one residual block since all of the other blocks has the same parameter. And we keep to this setting in Table~\ref{tbl:super_resolution_srresnet_edsr} and Table~\ref{tbl:super_resolution_edsr}. By default, each convolution layer uses an unique basis set, \ie, without basis sharing. In Table~\ref{tbl:super_resolution_edsr_ablation} and Table~\ref{tbl:super_resolution_edsr}, the corresponding basis sharing result is also shown. 

In Table~\ref{tbl:super_resolution_edsr_ablation}, there are several noticeable points. Firstly, the compressed models with and without basis sharing technique achieve almost the same PSNR for different $m$. But with basis sharing, the model size is further compressed. Secondly, when $m=64$ and basis sharing is used, the compressed model only accounts for 9\% of the parameters of the original network. When basis sharing is further used, an impressive compression rate of 5.9\% is achieved while the PSNR result is not far away from the baseline. Thirdly, the most aggressive compression ratio is 1.5\%. Considering that there are 32 and 16 residual blocks in EDSR and SRResNet respectively, this operating point brings the model size from EDSR level to SRResNet level while the PSNR of the resulting model is higher than that of SRResNet. 

In Table~\ref{tbl:super_resolution_srresnet_edsr}, the results of Factor~\cite{wang2017factorized} and our basis learning method are shown for the SRResNet and EDSR-8-128. A lighter and a heavier operating point are reported for each compression method. The proposed method outperforms Factor under the two settings. To further compare Factor and the proposed method, we apply the compression method to the fully-fledged EDSR model. As shown in Table~\ref{tbl:super_resolution_edsr}, the compressed model Basis-S is much better than Factor and in the meanwhile with fewer parameters. The PSNR result of our Basis-S is slightly worse than that of Basis. The super-resolved \textit{bird} images of compressed EDSR by different methods are shown in Fig~.\ref{fig:sr_bird}. The image from our compressed model has the highest PSNR and it is very close to the baseline in visual quality as well.
\subsection{Validation on image classification}
\label{subsec:classification}

\begin{table}[t]
    \small
    \begin{center}
        \begin{tabular}{c|c|c|c}
        \toprule
        Configuration & Top-1 Error (\%) & \#Params & Comp.(\%) \\ \midrule
        M24 & 5.69 &320k &30.8\\ 
        M26 & 5.70 &336k &32.3\\
        M32 & 5.57 &383k &36.8\\
        M36T6 & 5.32 &331k &31.8\\
        M38T12 & 5.56 &326k &31.3\\ \midrule
        Baseline & 5.26 &1041k &100 \\
        \bottomrule
        \end{tabular}
    \end{center}
    \vspace{-0.1cm}
    \caption{Different operating points of our method for compressing DenseNet-12-40 architecture~\cite{huang2017densely} for image classification. `M*' and `T*' means the number of basis in DenseBlock and the splits in the transition block in DenseNet, respectively. The classification error is Top-1 error for CIFAR10.}
    \vspace{-0.4cm}
    \label{tbl:classification_result_densenet}
\end{table}

\begin{table*}[t]
    \small
    \begin{center}
        \begin{tabular}{c|c|c|c|c}
        \toprule
        Model & Method & Top-1 Error (\%) / Baseline & \#Parameters & Compression Rate(\%) \\ \midrule
        \multirow{4}{*}{VGG-16} & K-means~\cite{son2018clustering} &6.24 / 5.98 &3.27M &22.2\\
        & Factor~\cite{wang2017factorized} &7.12 / 5.98 &3.34M &22.7 \\
        & Group~\cite{peng2018extreme} &6.69 / 5.98 &3.80M &25.9\\
        & \textbf{Basis (ours)} &\textbf{6.18} / 5.98 &3.21M & \textbf{21.8}\\ \midrule
        \multirow{7}{*}{DenseNet-12-40} & K-means &5.44 / 5.26 &335k &32.2\\
        & Factor &6.71 /5.26 & 317k & 30.4\\
        & Group &6.65 / 5.26 &337k &32.4\\
        & KSE~\cite{Li_2019_CVPR_KSE} &5.30 / 5.19 &390k &37.5\\
        & \cite{Liu_2017_ICCV_Learning}(70\%) &5.65 / 5.19 &350k &33.6\\
        & Simple-SVD &7.14 / 5.26 &360k &34.6\\
        & \textbf{Basis (ours)} &\textbf{5.32} / 5.26 &331k &31.8\\ \midrule
        \multirow{4}{*}{ResNet-56} & K-means~\cite{son2018clustering} &6.76 / 6.28 & 190k &22.4\\
        & Factor &8.70 / 6.28 &212k &24.9\\
        & Group &6.45 / 6.28 &206k &24.3\\
        & KSE &7.12 / 6.97 &360k &42.4\\
        & \textbf{Basis (ours)} &6.60 / 6.28 &186k &\textbf{21.9}\\
        \bottomrule
        \end{tabular}
    \end{center}
    \vspace{-0.15cm}
    \caption{Compression results for VGG~\cite{simonyan2014very}, DenseNet~\cite{huang2017densely} and ResNet~\cite{he2016deep} trained on CIFAR-10~\cite{krizhevsky2009learning}. For a fair comparison, the model size from different methods is kept to the same level.}
    \label{tbl:classification_all}
    \vspace{-0.4cm}
\end{table*}

\begin{figure}[t]
    \begin{center}
        \includegraphics[width=0.75\linewidth]{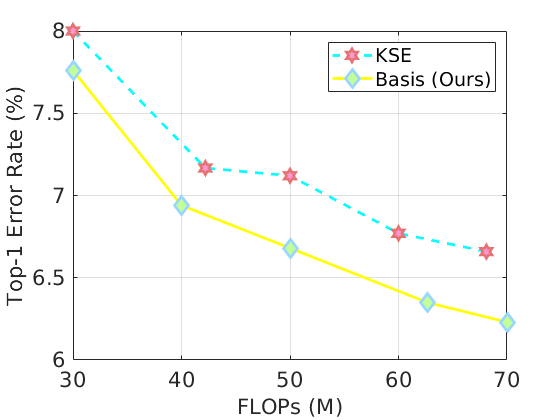}
    \end{center}
    \vspace{-0.2cm}
    \caption{Comparison between our method and KSE~\cite{Li_2019_CVPR_KSE} for ResNet-56 on CIFAR10.}
    \vspace{-0.4cm}
    \label{fig:comparison_basis_kse}
\end{figure}

\begin{table}[t]
    \footnotesize
    \begin{center}
        \begin{tabular}{c|c|c|c}
        \toprule
        Model & Method &Top-1 Err.(\%) / Baseline & FLOPs (\%) \\ \midrule
        \multirow{4}{*}{VGG} & K-means~\cite{son2018clustering} & 6.24 / 5.98 & 100 \\ 
        & Factor~\cite{wang2017factorized} &7.12 / 5.98 & 36.6 \\ 
        & Group~\cite{peng2018extreme} &6.69 / 5.98 & 46.1 \\ 
        & Basis (ours) &\textbf{6.23} / 5.98 & \textbf{23.5} \\ \midrule
        \multirow{5}{*}{ResNet56} & CaP~\cite{Minnehan_2019_CVPR_CaP} &6.78 / 6.49 & 50.2 \\
        & ENC~\cite{Kim_2019_CVPR_ENC} &7.00 / 6.90 & 50.0 \\
        & AMC~\cite{He_2018_ECCV_AMC} &8.10 / 7.20 & 50.0 \\
        & KSE~\cite{Li_2019_CVPR_KSE} &6.77 / 6.97 & 48.0 \\
        & Basis (ours) & \textbf{6.08} / 7.05 & 50.0\\
        \bottomrule
        \end{tabular}
    \end{center}
    \vspace{-0.15cm}
    \caption{Top-1 error vs. FLOPs reduction rate for VGG-16 and ResNet-56 on CIFAR10. For K-means, the practical FLOPs in the authors' code rather than the theoretical is reported.}
    \vspace{-0.4cm}
    \label{tbl:classification_result_vgg_resnet56}
\end{table}

In Table~\ref{tbl:classification_result_densenet}, we show different operating points for the proposed method applied on DenseNet-12-40 architecture. When the basis size increases from 24 to 32, the corresponding error rate decreases from 5.69\% to 5.57\%. In addition, by applying compression to the $1 \times 1$ convolution in the transition layer, we can save some parameter budget for the DenseBlock, which is relatively more important in the network. Thus, for `M36T6' and `M38T12', we can utilize more basis and at the same time with smaller number of parameters. Compared with `M32', `M36T6' further reduces the error rate by 0.25\%. Interestingly, although our `M38T12' model uses two more basis than `M36T6', the error rate rises a little bit. This is because `M38T12' uses an aggressive compression, \ie, $s=12$ in the transition block. Therefore, the compression degree of the DenseBlock and the transition block should be balanced to obtain the best trade-off between compression ratio and accuracy.

The compression results of different methods for VGG-16, DenseNet-12-40, and ResNet-56 are shown in Table~\ref{tbl:classification_all}. For a fair comparison, we follow the setting in~\cite{son2018clustering} for VGG-16. That is, only one instead of three fully-connected layer is appended after the last pooling layer. On VGG-16, our method shows the most aggressive compression and the lowest error rate. For our compressed model, we only suffer 0.2\% increase of error rate, which is quite small compared with 0.71\% 1.14\% increase of Group~\cite{peng2018extreme} and Factor. And our model has the smallest size. 
Our compression method and KSE~\cite{Li_2019_CVPR_KSE} shoots the lowest error rate on DenseNet-12-40. As for the compression ratio, although Factor is slightly lower than ours, its accuracy is the worst among all the compared methods. For ResNet-56, our method performs comparable with Group in terms of accuracy while with 20k fewer parameters. Table~\ref{tbl:classification_result_vgg_resnet56} and Fig.~\ref{fig:comparison_basis_kse} compare the computational cost of the proposed method and the state-of-the-art. Results are reported at operating points different from those in Table~\ref{tbl:classification_all}. For VGG-16, our method achieves the lowest error rate under the severest FLOPs reduction. For ResNet-56, the proposed method outperforms the others by a significant margin under the same FLOPs reduction. In Fig.~\ref{fig:comparison_basis_kse}, our method always shoots a lower error rate than KSE.

\section{Conclusion}
\label{sec:conclusion}
In this paper, we explored how to learn a filter basis set for convolution operations in modern CNNs. Our method is not limited by the filter size. Thus, it can be applied to $1 \times 1$ convolution and convolution with large input channel and smaller output channel. We applied our basis learning method to image classification and SR networks. The experiments validate the advantage of our basis learning method. Our compressed SRResNet and EDSR outperforms the models from the previous filter decomposition method. For the image SR network EDSR, the most aggressive compression our method brings the model size from EDSR level to SRResNet level while being more accurate than SRResNet. For VGG-16, the error rate of the model compressed by our method is within 0.2\% the baseline error, which is much better than the results of other compression methods. Our filter basis learning method leads to state-of-the-art performance on ResNet and DenseNet.

{\noindent\textbf{Acknowledgments:} This work was supported by Huawei, the ETH Zurich General Fund, and an Nvidia GPU hardware grant.
}

{\small
\bibliographystyle{ieee_fullname}
\bibliography{ms}
}
\appendix
\onecolumn
\clearpage
\section{Compressed Network Configuration}
\label{sec:configuration}

The basis configurations of our filter basis learning method for different networks including DenseNet~\cite{huang2017densely}, ResNet~\cite{he2016deep}, VGG~\cite{simonyan2014very}, EDSR~\cite{lim2017enhanced}, EDSR-8-128, SRResNet~\cite{ledig2016photo} are shown in Table~\ref{tbl:configuration_densent}, Table~\ref{tbl:configuration_resnet}, and Table~\ref{tbl:configuration_all}. For DenseNet, we used the network-wise basis sharing. For ResNet, we used group-wise basis sharing. We also tried basis sharing within the residual block for EDSR.

We reimplemented the network compression method Factor~\cite{wang2017factorized} and Group~\cite{peng2018extreme}. For the Factor method, to compare the methods fairly, we use two and three single intra-channel convolutional layers (SIC layer)~\cite{wang2017factorized} in Table~\ref{tbl:super_resolution_srresnet_edsr}, two SIC layers in Table~\ref{tbl:classification_all} and Table~\ref{tbl:classification_result_vgg_resnet56}, and one SIC layer in Table~\ref{tbl:super_resolution_edsr} to substitute one standard convolutional layer. To keep the number of parameter of the Group method~\cite{peng2018extreme} at the same level with other methods, the group size is set to 8, and 64 to approximate ResNet, and VGG respectively. To compress DenseNet, 3 groups are used for the first 20 DenseBlocks while 6 groups are used for the rest DenseBlocks. 

\begin{table}[b]
    \footnotesize
    \begin{center}
        \begin{tabular}{c|c|c}
            \toprule
            Number of basis $m$ & Channel $c$ &Transition Layer Split \\ \midrule
            24 & 444 & 1 \\ \hline
            26 & 444 & 1 \\ \hline
            32 & 444 & 1 \\ \hline
            36 & 444 & 6 \\ \hline
            38 & 444 & 12 \\
            \bottomrule
        \end{tabular}
    \end{center}
    \caption{DenseNet-12-40 compression configuration for Table~\ref{tbl:classification_result_densenet} in the main paper. The basis set is shared by all of the DenseBlocks. For lower layer DenseBlocks, a slice of the shared basis is used as the basis of that layer. For the former three configurations, we do not compress the transition layers in DenseNet. But for the latter two, the transition layers are also compressed with the specified number of splits.}
    \label{tbl:configuration_densent}
\end{table}

\begin{figure*}[b]
\begin{center}
   \includegraphics[width=0.8\linewidth]{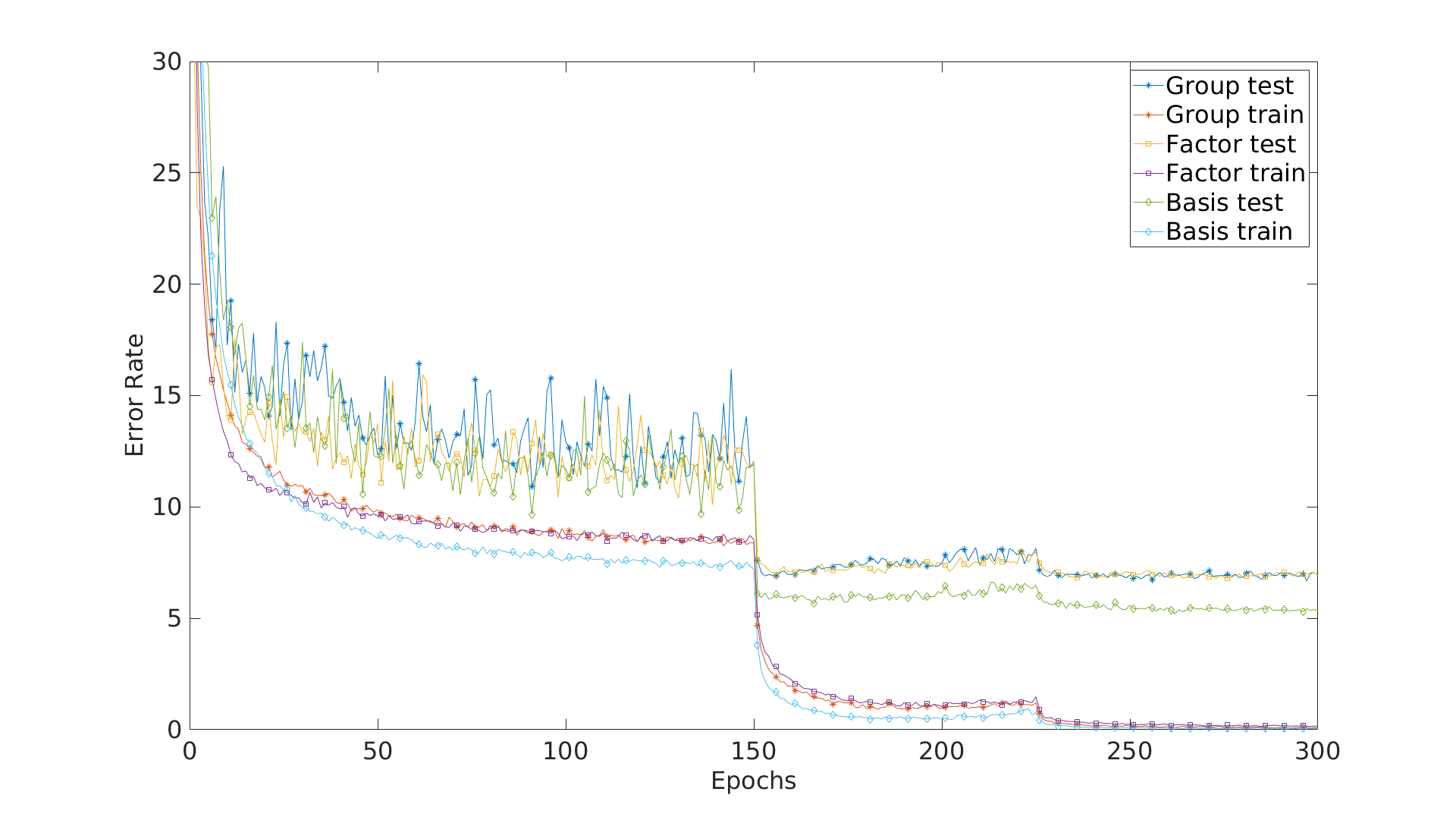}
\end{center}
   \caption{Training and testing error of different compression method applied on DenseNet-12-40.}
\label{fig:test_train_log_densenet}
\end{figure*}

\begin{figure*}[t]
\begin{center}
   \includegraphics[width=0.8\linewidth]{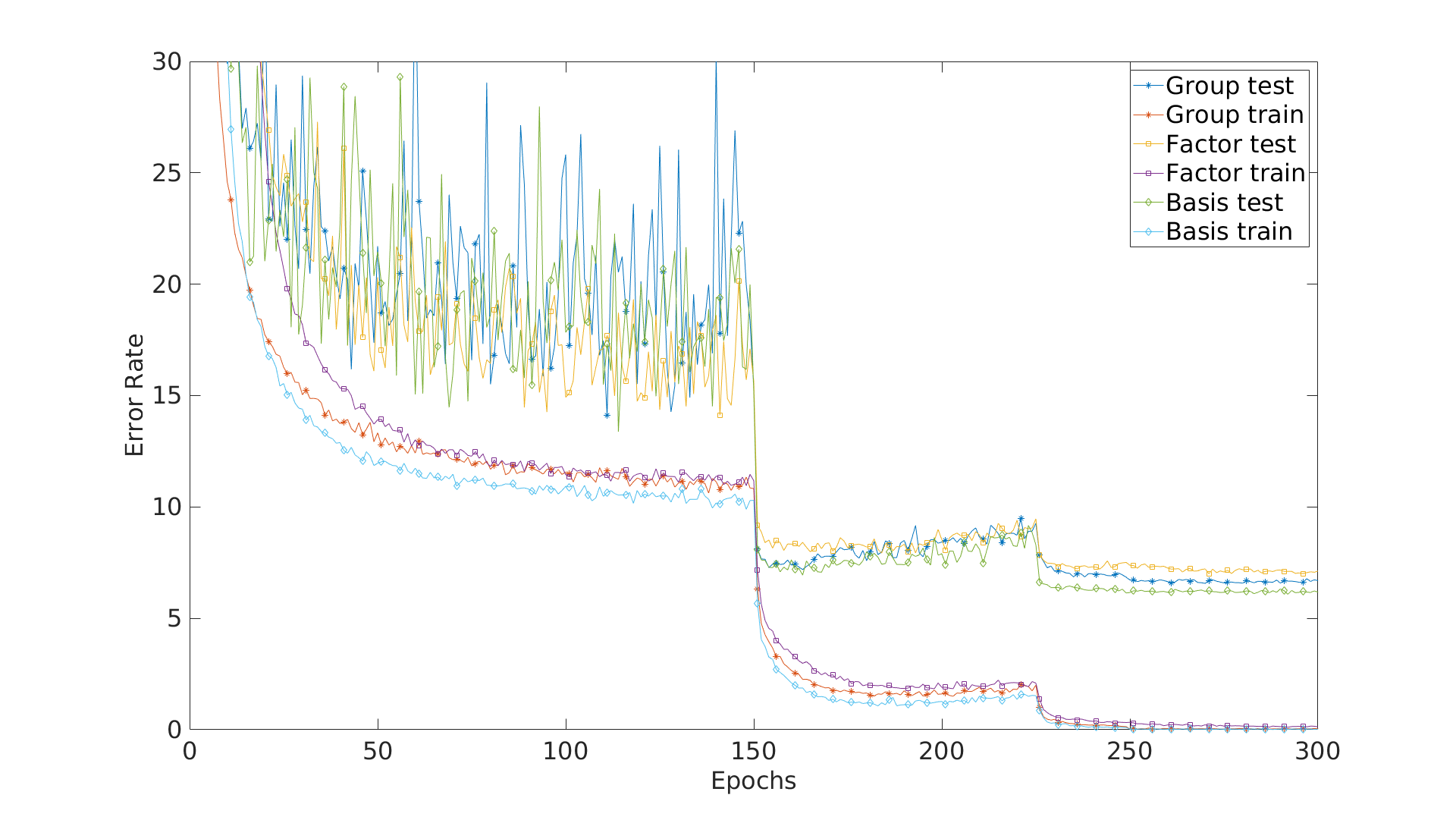}
\end{center}
   \caption{Training and testing error of different compression method applied on VGG-16.}
\label{fig:test_train_log_vgg}
\end{figure*}

\begin{figure*}[t]
\begin{center}
   \includegraphics[width=0.8\linewidth]{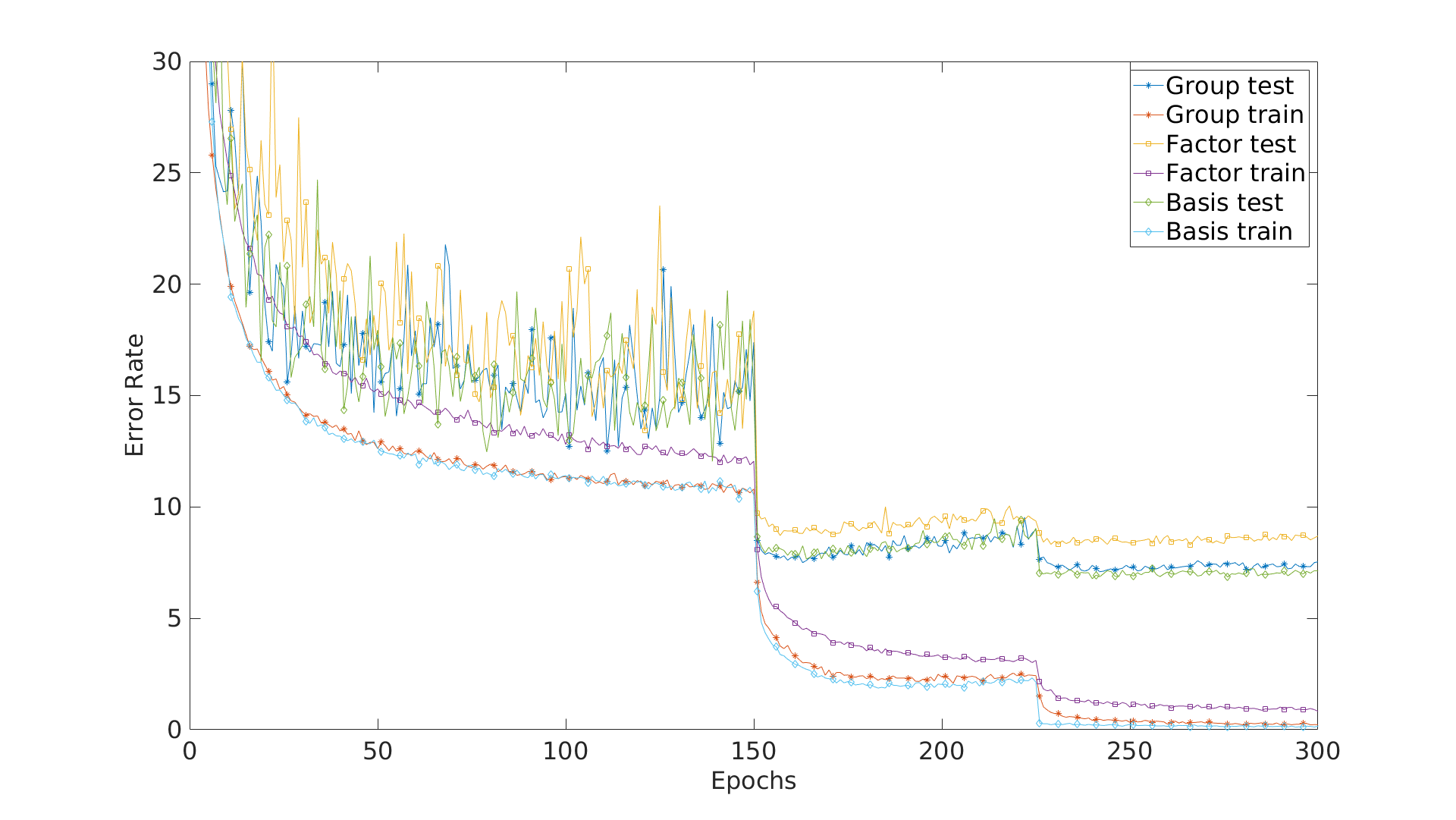}
\end{center}
   \caption{Training and testing error of different compression method applied on ResNet-56.}
\label{fig:test_train_log_resnet}
\end{figure*}

\begin{table}[t]
    \footnotesize
    \begin{center}
        \begin{tabular}{c|c|c}
            \toprule
            Residual Block Group & Number of basis $m$ & Channel $c$ \\ \midrule
            Group One & 24 & 16 \\ \hline
            Group Two & 48 & 32 \\ \hline
            Group Three & 84 & 64 \\
            \bottomrule
        \end{tabular}
    \end{center}
    \caption{ResNet-56 compression configuration. There are 27 residual blocks in ResNet-56, distributed into three groups with increasing number of channels but reducing resolution. The basis is shared by the convolutions within the same group. This configuration corresponds to the ResNet-56 entry in Table~\ref{tbl:classification_all} of the main paper.}
    \label{tbl:configuration_resnet}
\end{table}

\begin{table}[t]
    \footnotesize
    \begin{center}
        \begin{tabular}{c|c|c}
            \toprule
            Network & Number of basis $m$ & Channel $c$ \\ \midrule
            SRResNet (Basis-64-14) & 14 & 64 \\
            SRResNet (Basis-32-32) & 32 & 32 \\ \hline
            EDSR-8-128 (Basis-128-27) & 27 & 128 \\ 
            EDSR-8-128 (Basis-128-40) & 40 & 128 \\ \hline
            EDSR (Basis) & 32 & 256 \\ \hline
            EDSR (Basis-S) & 32 & 256 \\ \hline
            VGG-16 & 128 & 128 \\
            \bottomrule
        \end{tabular}
    \end{center}
    \caption{Compression configuration of SRResNet, EDSR-8-128, EDSR, and VGG-16. `Basis' means that there is a unique basis for each convolutional layer. `Basis-S' means that the basis is shared by the two convolutional layers within the residual block. For VGG-16, the first three convolutional layers are not compressed.}
    \label{tbl:configuration_all}
\end{table}

\begin{figure*}
\centering
\subfigure{
  \begin{minipage}[t]{0.19\linewidth}
    \centering
    \includegraphics[width=1\textwidth]{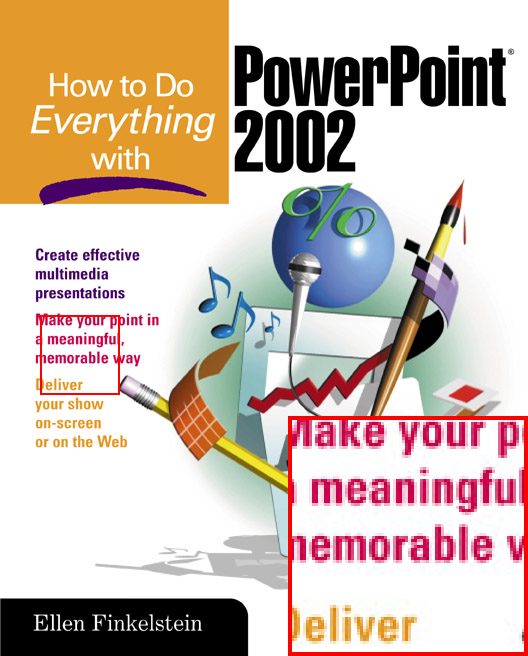}
    {\footnotesize Ground-Truth: PSNR (dB)
    }
  \end{minipage}
  
  \begin{minipage}[t]{0.19\linewidth}
    \centering
    \includegraphics[width=1\textwidth]{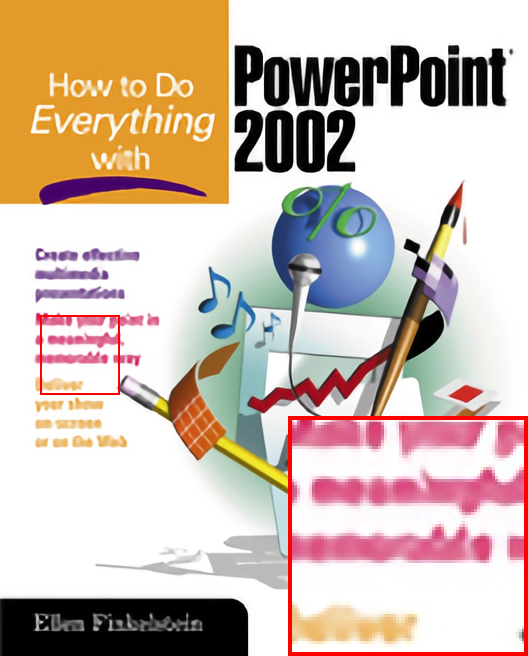}
    {\footnotesize Factor: 26.70 dB
    }
  \end{minipage}
  \begin{minipage}[t]{0.19\linewidth}
    \centering
    \includegraphics[width=1\textwidth]{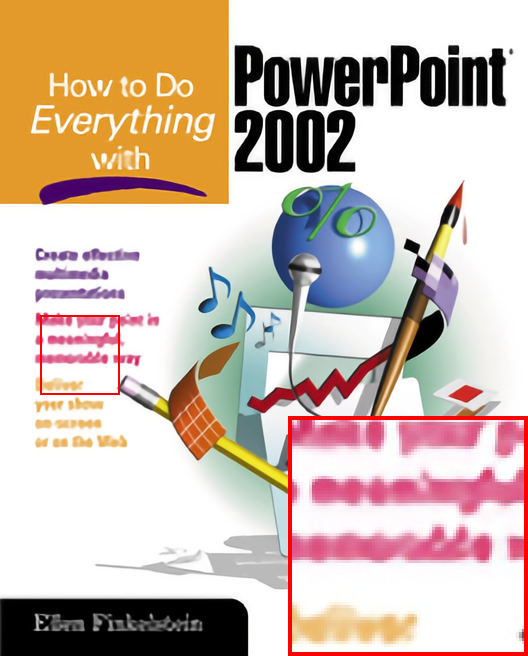}
    {\footnotesize Basis-S (ours): 26.79 dB
    }
  \end{minipage}
  \begin{minipage}[t]{0.19\linewidth}
    \centering
    \includegraphics[width=1\textwidth]{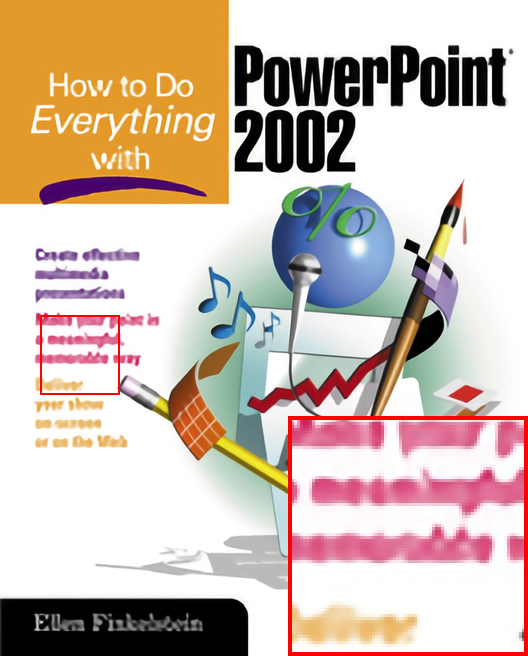}
    {\footnotesize Basis (ours): 27.01 dB}
  \end{minipage}
  \begin{minipage}[t]{0.19\linewidth}
    \centering
    \includegraphics[width=1\textwidth]{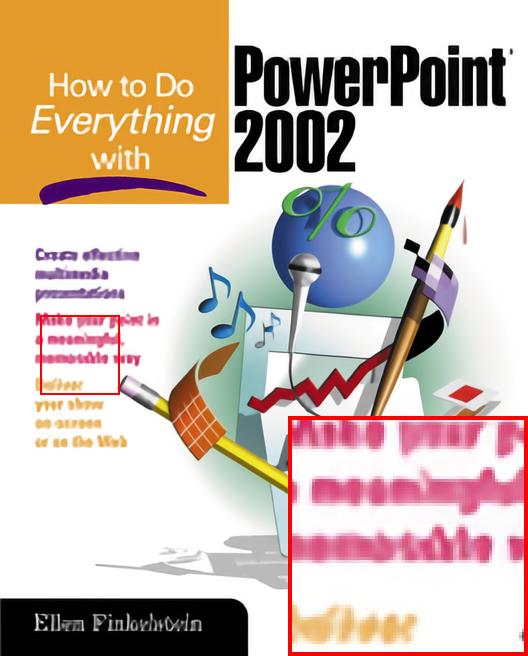}
    {\footnotesize Baseline: 27.17 dB}
  \end{minipage}
}
\subfigure{
     \begin{minipage}[t]{0.19\linewidth}
    \centering
    \includegraphics[width=1\textwidth]{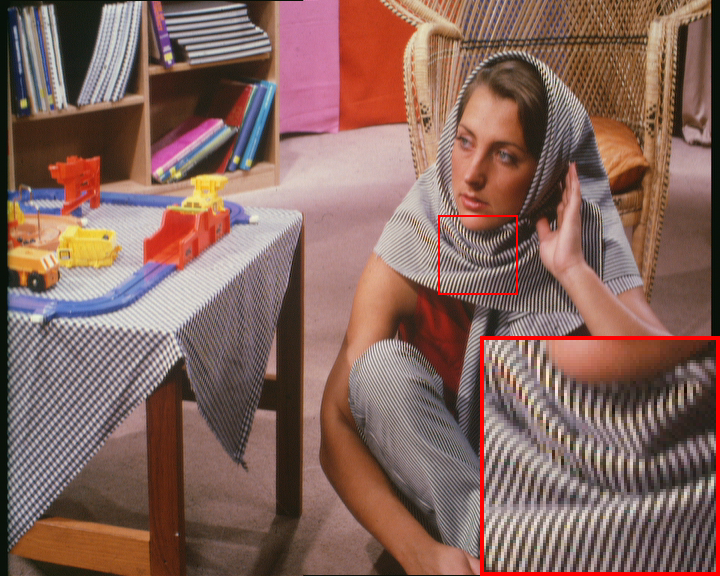}
    {\footnotesize Ground-Truth: PSNR (dB)
    }
  \end{minipage}
  
  \begin{minipage}[t]{0.19\linewidth}
    \centering
    \includegraphics[width=1\textwidth]{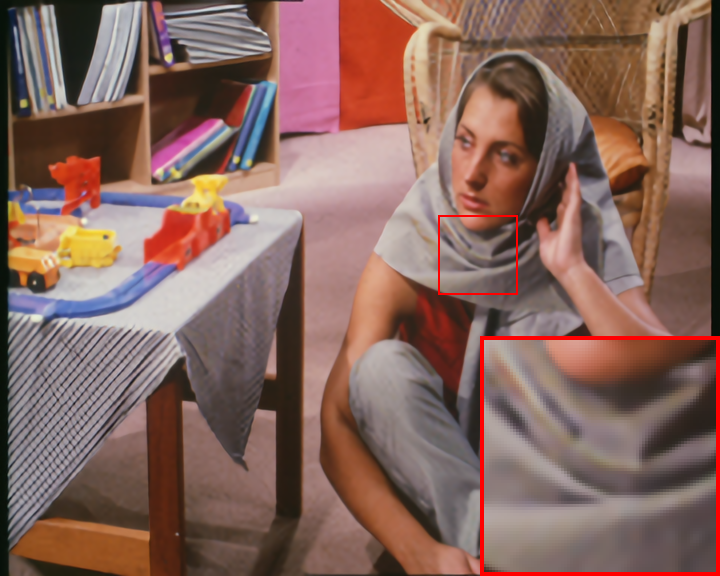}
    {\footnotesize Factor: 25.86 dB
    }
  \end{minipage}
  \begin{minipage}[t]{0.19\linewidth}
    \centering
    \includegraphics[width=1\textwidth]{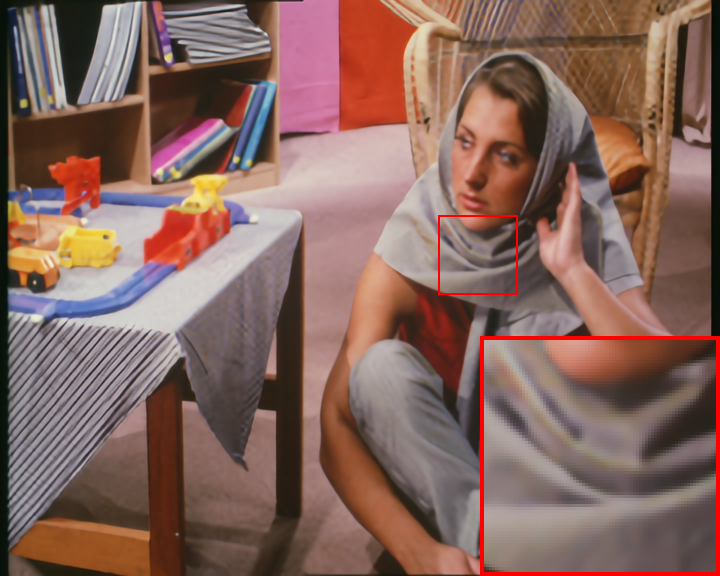}
    {\footnotesize Basis-S (ours): 25.94 dB
    }
  \end{minipage}
  \begin{minipage}[t]{0.19\linewidth}
    \centering
    \includegraphics[width=1\textwidth]{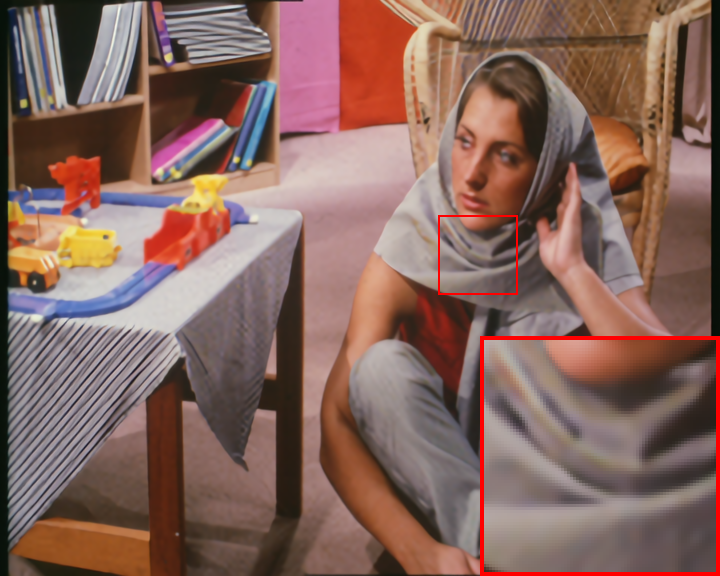}
    {\footnotesize Basis (ours): 25.96 dB}
  \end{minipage}
  \begin{minipage}[t]{0.19\linewidth}
    \centering
    \includegraphics[width=1\textwidth]{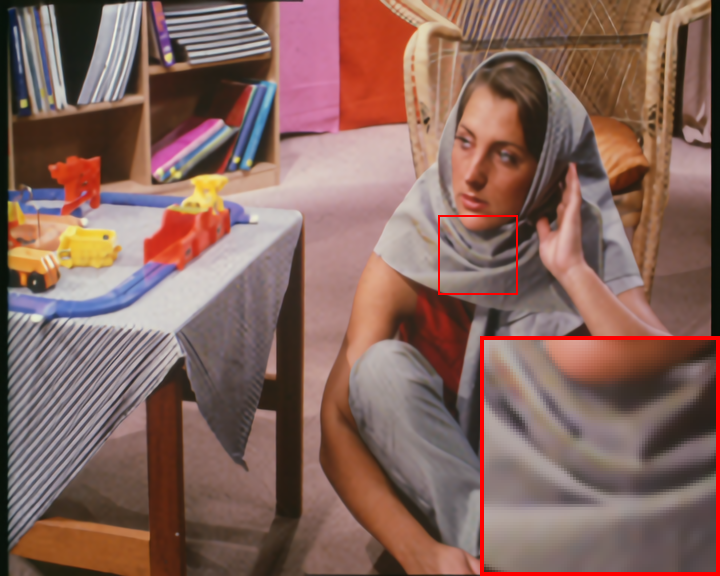}
    {\footnotesize Baseline: 26.45 dB}
  \end{minipage}
}
\subfigure{
     \begin{minipage}[t]{0.19\linewidth}
    \centering
    \includegraphics[width=1\textwidth]{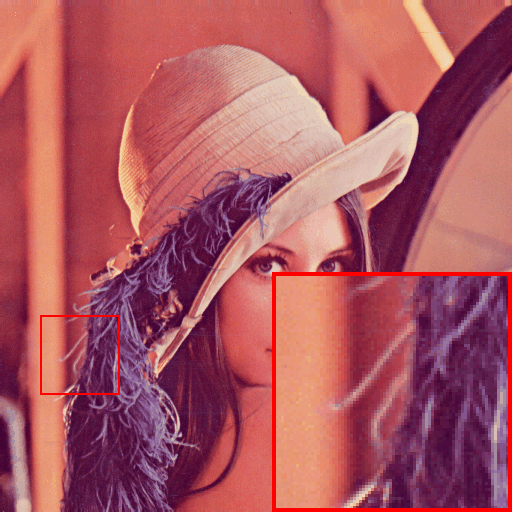}
    {\footnotesize Ground-Truth: PSNR (dB)
    }
  \end{minipage}
  
  \begin{minipage}[t]{0.19\linewidth}
    \centering
    \includegraphics[width=1\textwidth]{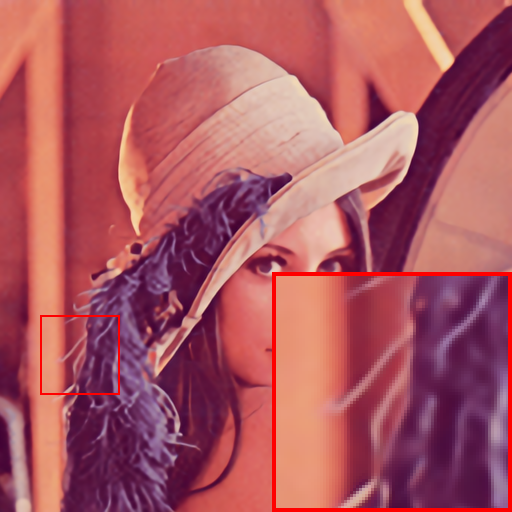}
    {\footnotesize Factor: 32.47 dB
    }
  \end{minipage}
  \begin{minipage}[t]{0.19\linewidth}
    \centering
    \includegraphics[width=1\textwidth]{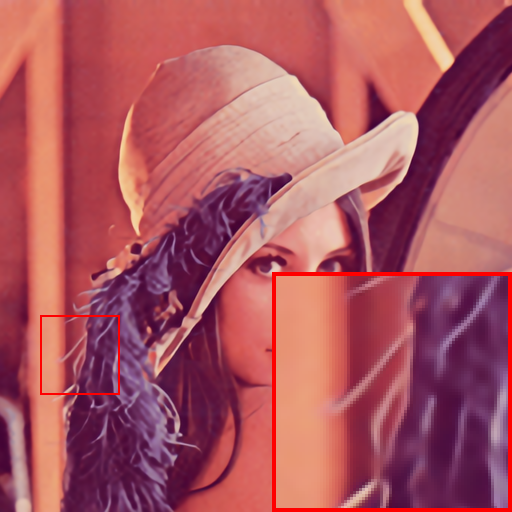}
    {\footnotesize Basis-S (ours): 32.48 dB
    }
  \end{minipage}
  \begin{minipage}[t]{0.19\linewidth}
    \centering
    \includegraphics[width=1\textwidth]{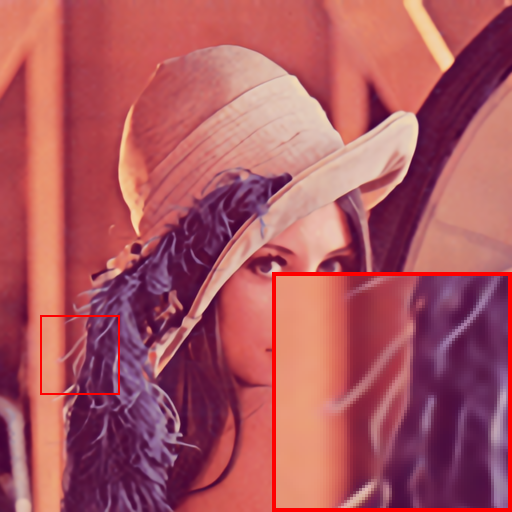}
    {\footnotesize Basis (ours): 32.68 dB}
  \end{minipage}
  \begin{minipage}[t]{0.19\linewidth}
    \centering
    \includegraphics[width=1\textwidth]{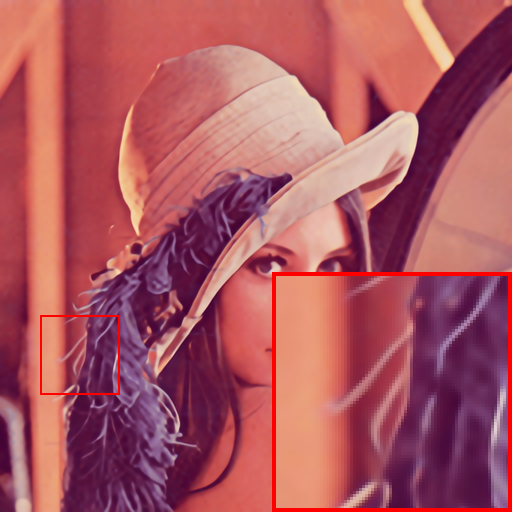}
    {\footnotesize Baseline: 32.69 dB}
  \end{minipage}
}
\caption{SR results for upscaling factor $\times 4$. Network compression methods are applied on EDSR. PSNR values are reported.} 
\label{fig:edsr_8_128}
\end{figure*}

\section{Training and Testing Error Curves for Image Classification}
\label{sec:curves}

The error curves during training and testing for DenseNet-12-40, ResNet-56, and VGG-16 on CIFAR10 are shown in Fig.~\ref{fig:test_train_log_densenet}, Fig~\ref{fig:test_train_log_resnet}, and Fig~\ref{fig:test_train_log_vgg}, respectively. Our method shoots the lowest stable error rate for all the three networks during training and testing. 

\section{More Visual Results for Super-Resolution}

\begin{figure*}
\centering
\subfigure{
  \begin{minipage}[t]{0.19\linewidth}
    \centering
    \includegraphics[width=1\textwidth]{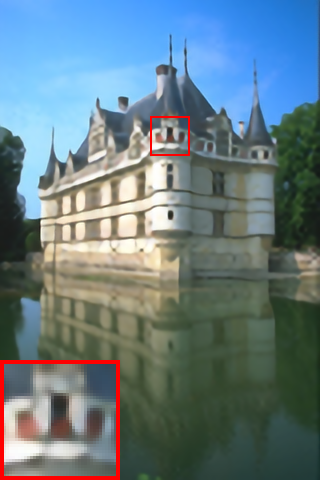}
    {\footnotesize Factor-SIC2: 26.44 dB
    }
  \end{minipage}
  \begin{minipage}[t]{0.19\linewidth}
    \centering
    \includegraphics[width=1\textwidth]{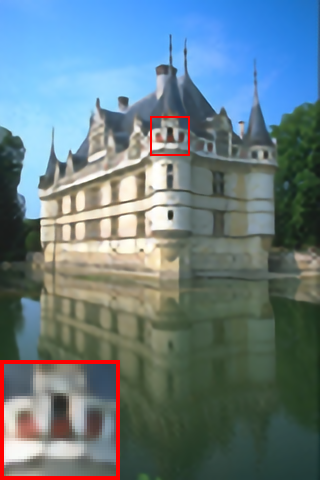}
    {\footnotesize Factor-SIC3: 26.47 dB
    }
  \end{minipage}
  \begin{minipage}[t]{0.19\linewidth}
    \centering
    \includegraphics[width=1\textwidth]{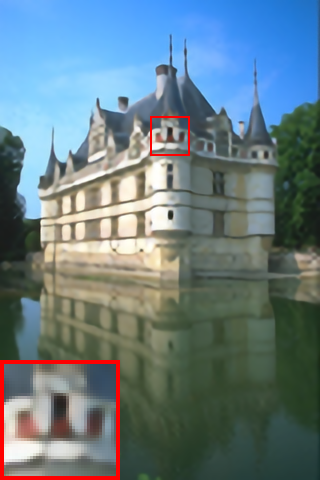}
    {\footnotesize Basis-14-64 (ours): 26.47 dB}
  \end{minipage}
  \begin{minipage}[t]{0.19\linewidth}
    \centering
    \includegraphics[width=1\textwidth]{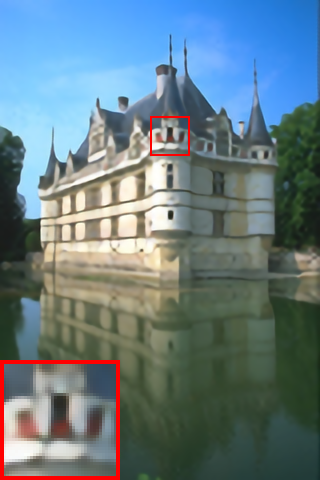}
    {\footnotesize Basis-32-32 (ours): 26.57 dB}
  \end{minipage}
    \begin{minipage}[t]{0.19\linewidth}
    \centering
    \includegraphics[width=1\textwidth]{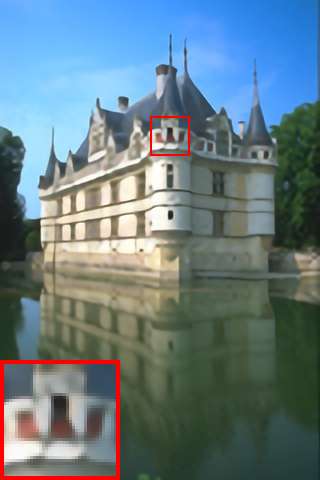}
    {\footnotesize Baseline: 26.65 dB}
  \end{minipage}
}
\subfigure{
 
  \begin{minipage}[t]{0.19\linewidth}
    \centering
    \includegraphics[width=1\textwidth]{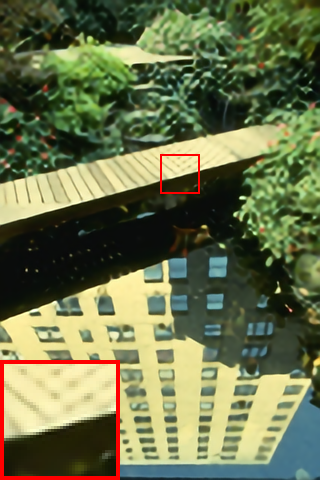}
    {\footnotesize Factor-SIC2: 21.88 dB
    }
  \end{minipage}
  \begin{minipage}[t]{0.19\linewidth}
    \centering
    \includegraphics[width=1\textwidth]{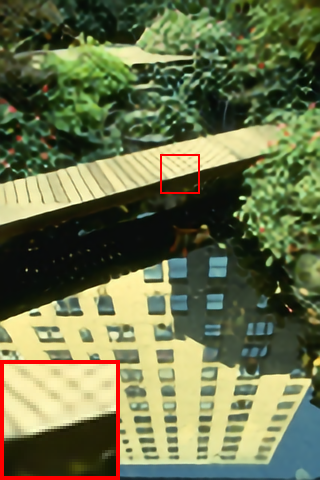}
    {\footnotesize Factor-SIC3: 21.91 dB
    }
  \end{minipage}
  \begin{minipage}[t]{0.19\linewidth}
    \centering
    \includegraphics[width=1\textwidth]{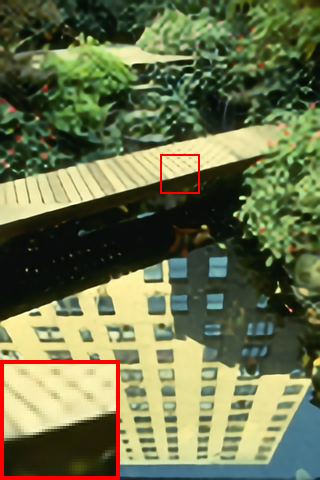}
    {\footnotesize Basis-14-64 (ours): 21.91 dB}
  \end{minipage}
  \begin{minipage}[t]{0.19\linewidth}
    \centering
    \includegraphics[width=1\textwidth]{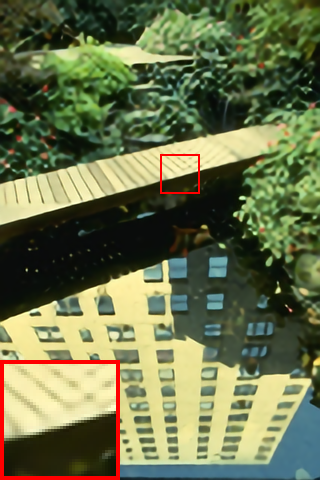}
    {\footnotesize Basis-32-32 (ours): 22.01 dB}
  \end{minipage}
    \begin{minipage}[t]{0.19\linewidth}
    \centering
    \includegraphics[width=1\textwidth]{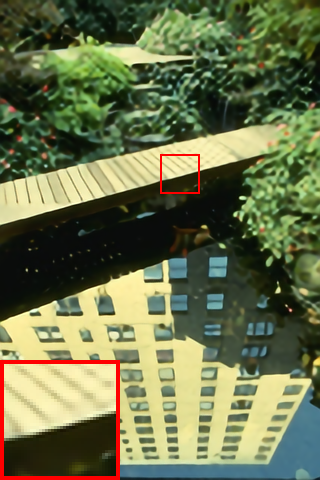}
    {\footnotesize Baseline: 22.09 dB}
  \end{minipage}
}
\subfigure{
 
  \begin{minipage}[t]{0.19\linewidth}
    \centering
    \includegraphics[width=1\textwidth]{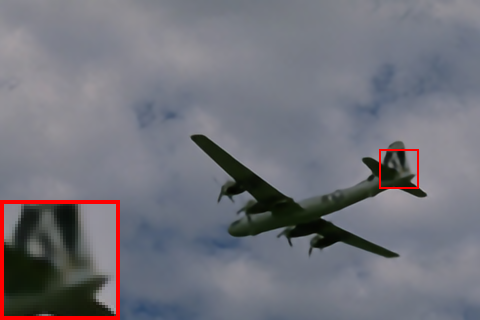}
    {\footnotesize Factor-SIC2: 39.68 dB
    }
  \end{minipage}
  \begin{minipage}[t]{0.19\linewidth}
    \centering
    \includegraphics[width=1\textwidth]{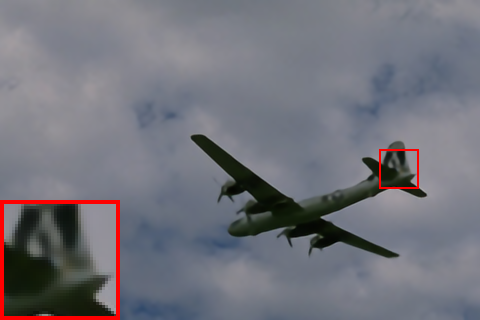}
    {\footnotesize Factor-SIC3: 39.80 dB
    }
  \end{minipage}
  \begin{minipage}[t]{0.19\linewidth}
    \centering
    \includegraphics[width=1\textwidth]{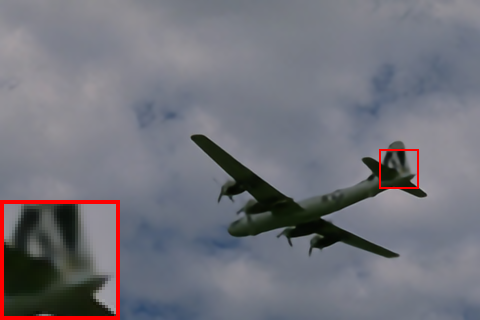}
    {\footnotesize Basis-14-64 (ours): 39.74 dB}
  \end{minipage}
  \begin{minipage}[t]{0.19\linewidth}
    \centering
    \includegraphics[width=1\textwidth]{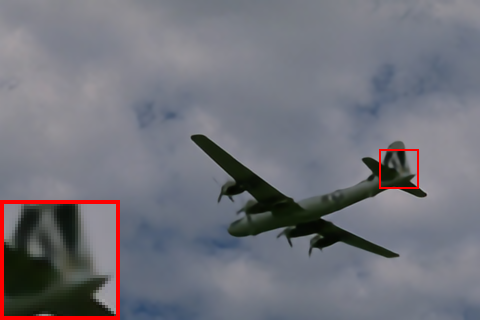}
    {\footnotesize Basis-32-32 (ours): 39.94 dB}
  \end{minipage}
    \begin{minipage}[t]{0.19\linewidth}
    \centering
    \includegraphics[width=1\textwidth]{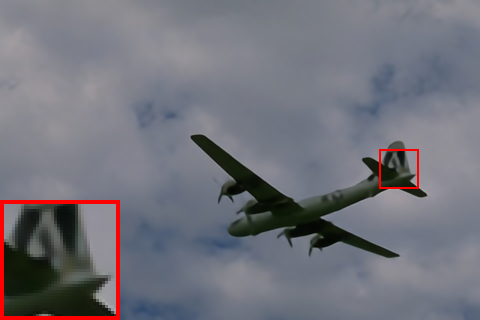}
    {\footnotesize Baseline: 40.28 dB}
  \end{minipage}
}
\caption{SR results for upscaling factor $\times 4$. Network compression methods are applied on SRResNet. PSNR values are reported.} 
\label{fig:srresnet}
\end{figure*}

More visual results for image super-resolution are shown in Fig.~\ref{fig:srresnet} and Fig.~\ref{fig:edsr_8_128} for compressing SRResNet and EDSR-8-128 respectively. Compared with Factor~\cite{wang2017factorized} and Group~\cite{peng2018extreme}, the SR images from our compressed model are very close to the baseline in terms of both visual quality and PSNR values.

\end{document}